\documentclass[runningheads]{llncs}

\usepackage{eccv}

\usepackage{eccvabbrv}

\usepackage{graphicx}
\usepackage{booktabs}
\usepackage[dvipsnames]{xcolor}
\definecolor{mplblue}{HTML}{1F77B4}
\definecolor{mplred}{HTML}{D62728}
\definecolor{mplgreen}{HTML}{2CA02C}
\usepackage{tabularx}
\usepackage{siunitx}
\usepackage{tikz}
\usepackage{algorithm}
\usepackage{algpseudocode}
\usepackage{amsmath}
\usepackage{makecell}
\usepackage{multirow}
\usepackage{pifont}
\usepackage{caption}
\usepackage{subcaption}
\usepackage{wrapfig}

\usepackage[normalem]{ulem}

\usepackage[accsupp]{axessibility}  

\newcommand*{\inparagraph}[1]{\noindent\textbf{#1}\hspace{0.5em}}

\newcommand \eq[1]{\begin{equation}\begin{aligned}#1\end{aligned}\end{equation}}
\newcommand*\diff{\mathop{}\!\mathrm{d}}

\newcommand{\pointcloud}{\mathcal{P}}
\newcommand{\invtemp}{\eta}
\newcommand{\guidancestrength}{\beta}
\newcommand{\latent}{z}
\newcommand{\point}{x}
\newcommand{\shape}{S}
\newcommand{\sep}{|}

\newcommand{\pguided}{\tilde p}

\makeatletter
\newcommand{\geomloss}{\@ifnextchar\bgroup\geomloss@withargs\geomloss@noargs}
\newcommand{\geomloss@noargs}{\mathcal{L}}
\newcommand{\geomloss@withargs}[2]{\mathcal{L}(#1,#2)}
\makeatother
\newcommand{\dec}{D}
\newcommand{\enc}{E}
\newcommand{\surfloss}{\mathcal{L}_\text{surface}}
\newcommand{\eikloss}{\mathcal{L}_\text{eikonal}}
\newcommand{\sirenloss}{\mathcal{L}_\text{siren}}

\newcommand{\emdash}{\,---\,}

\usepackage[pagebackref,breaklinks,colorlinks,citecolor=eccvblue]{hyperref}

\usepackage{orcidlink}

\begin{document}

\title{Generative Shape Reconstruction with Geometry-Guided Langevin Dynamics}
\titlerunning{Geometry-Guided Langevin Dynamics}
\author{
Linus Härenstam-Nielsen\inst{1,2}
\and Dmitrii Pozdeev\inst{1}
\and Thomas Dagès\inst{1,2}
\and\\ Nikita Araslanov\inst{1,2}
\and Daniel Cremers\inst{1,2}
}
\authorrunning{Härenstam-Nielsen \etal}
\institute{Technical University of Munich\\
\and
Munich Center for Machine Learning\\
}

\maketitle

\begin{abstract}
\sloppy
Reconstructing complete 3D shapes from incomplete or noisy observations is a fundamentally ill-posed problem that requires balancing measurement consistency with shape plausibility. Existing methods for shape reconstruction can achieve strong geometric fidelity in ideal conditions but fail under realistic conditions with incomplete measurements or noise. At the same time, recent generative models for 3D shapes can synthesize highly realistic and detailed shapes but fail to be consistent with observed measurements. In this work, we introduce GG-Langevin: Geometry-Guided Langevin dynamics, a probabilistic approach that unifies these complementary perspectives. By traversing the trajectories of Langevin dynamics induced by a diffusion model, while preserving measurement consistency at every step, we generatively reconstruct shapes that fit both the measurements and the data-informed prior. We demonstrate through extensive experiments that GG-Langevin achieves higher geometric accuracy and greater robustness to missing data than existing methods for surface reconstruction. 

\end{abstract}

\begin{figure}[t]
\centering 
\includegraphics[width=\textwidth,trim=75 30 30 10,clip]{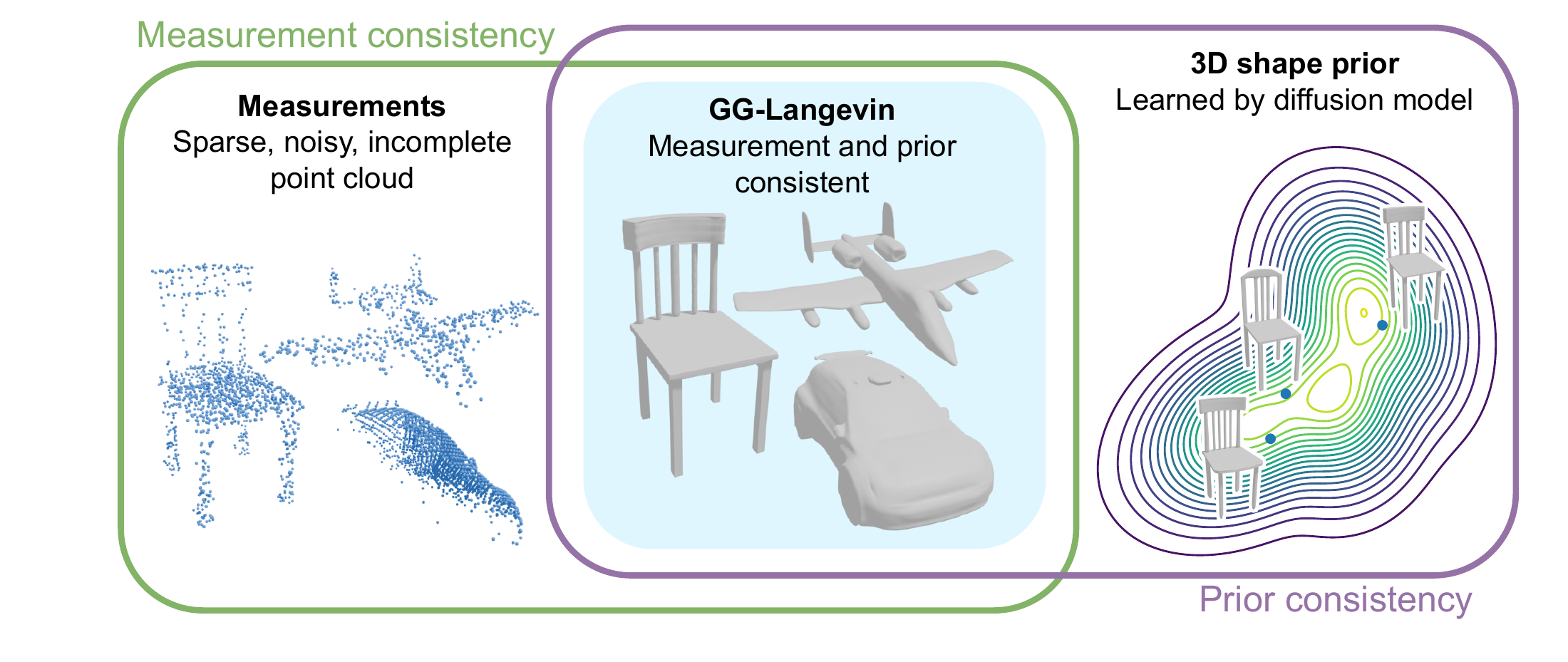}
\caption{%
    \textbf{GG-Langevin.} We combine the prior learned by a diffusion model with gradients from a geometric loss at inference time. By guiding the trajectories of Langevin dynamics, we obtain shapes that are both measurement-consistent and prior-consistent.
}
\vspace{-0.8cm}
\label{fig:teaser1}
\end{figure}

\label{sec:introduction}
\section{Introduction}
\begingroup
\renewcommand{\thefootnote}{}
\footnotetext{Project available at \url{https://github.com/linusnie/gg-langevin}.}
\endgroup
Reconstructing complete shapes from incomplete point clouds is a central challenge in 3D reconstruction with applications in robotics, 3D scanning, and augmented reality.
Sensors, such as LiDAR and depth cameras, produce noisy, sparse, and incomplete point clouds, and the task is to recover a full surface that explains these observations.
The problem is inherently ambiguous: noise must be disambiguated from structure, and there are often multiple plausible shapes that explain the same input.
Solving this problem requires simultaneously enforcing measurement consistency (agreement with the observed geometry) and prior consistency (agreement with the manifold of realistic shapes).

Two dominant paradigms tackle these aspects independently.
Optimization-based methods, such as IGR~\cite{gropp2020implicit} and DiffCD~\cite{harenstam2024diffcd}, fit an implicit surface to the data by minimizing a geometric loss function. Optimization-based methods excel at enforcing measurement consistency but lack data-informed priors, leading to 
oversmoothed or implausible 
results when observations are missing or unreliable.
In contrast, learning-based approaches, such as NKSR~\cite{Huang:2023:NKS} and ShapeFormer~\cite{yan2022shapeformer}, learn to infer shapes directly from point clouds by training on large datasets of synthetically generated point cloud scans.
Yet in practice, these models often fail to simultaneously preserve both measurement consistency and prior consistency, especially when the noise model at inference time differs from the noise model used during training.

Separately from reconstruction, 3D generative models have advanced significantly in recent years, particularly diffusion and flow models.
These models provide highly accurate estimates of the prior distribution of 3D shapes when trained at sufficient scale, synthesizing detailed and realistic shapes.
However, while accurately capturing the prior, effectively leveraging generative models for 3D reconstruction remains an open problem.
Our work closes this critical gap.

In this work, we combine the benefits of optimization-based methods with the sample quality of generative models by leveraging a generative model as a prior.
By doing so, we simultaneously achieve both high measurement consistency and high prior consistency, as demonstrated in \cref{fig:teaser1}.
Our key insight is to reinterpret the optimization problem probabilistically as sampling shapes from a geometry-guided shape distribution.
We can then replace optimization trajectories with stochastic trajectories using Langevin dynamics, guided by the gradients of a geometric loss function.
In particular, we construct the geometry-guided shape distribution by weighting the prior distribution with a per-sample geometric weight, such that the sampling trajectories inherently lead to shapes that are both measurement-consistent and prior-consistent.
We refer to our approach as GG-Langevin (Geometry-Guided Langevin dynamics).

To efficiently sample from the geometry-guided distribution, we develop a novel Half-Denoising-No-Denoising (HDND) sampling algorithm,
which enables the diffusion model to operate on noisy latents (half-denoising) while the geometric loss operates on denoised latents (no-denoising).
Crucially, the half-denoising component relies on recent theory developed by Hyvärinen \cite{hyvarinen2024noise}, which we extend with guidance.
Furthermore, since our method operates in the latent space of a VAE, it repeatedly invokes the decoder during sampling, which necessitates an inexpensive yet accurate decoder.
To address this, we rebalance the architecture of the widely adopted VecSet-based VAE \cite{zhang20233dshape2vecset} by moving the encoder-decoder bottleneck, yielding a smaller decoder.
Interestingly, our rebalancing improves not only inference speed but also reconstruction quality.

We validate our approach by establishing two challenging surface reconstruction benchmarks with sparse and incomplete point clouds. 
In terms of reconstruction accuracy, GG-Langevin consistently outperforms prior state-of-the-art methods by a substantial margin across all tested object categories.
Our core contributions can be summarized as follows:

\begin{itemize}
    \item\textbf{GG-Langevin.}
    We combine neural implicit surface fitting with the generative prior from a pre-trained diffusion model, using Langevin dynamics as the theoretical basis.
    Our generative shape reconstruction method bridges the worlds of optimization and generative models, yielding highly accurate 3D shapes from sparse, noisy, and incomplete point clouds.
    
    \item \textbf{HDND.} We extend the recently developed half-denoising formulation \cite{hyvarinen2024noise} with denoising-free guidance.
    Our hybrid Half-Denoising-No-Denoising approach is particularly suited for the complex guidance functions typically employed for surface reconstruction.
    \item\textbf{Rebalanced shape VAE.}
    To enable efficient inference with GG-Langevin, we carefully rebalance the reference VecSet \cite{zhang20233dshape2vecset} VAE architecture by moving the bottleneck.
    We then train our diffusion model on the new latent space.
\end{itemize}

\section{Related work}
\subsection{Shape reconstruction}
Existing shape reconstruction approaches can be broadly categorized as follows: 
\textit{i)} optimization-based, where the shape is estimated by minimizing a hand-crafted loss function, \textit{ii)}~learning-based, where it is estimated with a feed-forward model trained on correspondences between measurements and full shapes, or \textit{iii)} optimization-based with a learned prior.

\inparagraph{Optimization-based.}
Optimization-based methods work by defining a loss function, which can be minimized iteratively \cite{Atzmon:2020:SAL, Ma2020NeuralPull, sitzmann2019siren, atzmon2021sald, lipman2021phase, harenstam2024diffcd, wang2025hotspot, coiffier2024LipschitzNDF, ling2025stochastic}.
Various works propose to regularize optimization-based methods with additional loss terms to stabilize training \cite{BenShabat2021DiGS, wang2023aligningHessianGradient, zixiong23neuralsingular, sitzmann2019siren, yang2023steik}, typically biasing the reconstruction toward smoother surfaces \cite{harenstam2024diffcd, yang2023steik}.
Due to their iterative nature, optimization-based methods can achieve strong consistency with the provided measurements.
However, due to the lack of a data-informed prior, these methods struggle with partial measurements or extreme noise.

\inparagraph{Learning-based.}
Another line of work estimates shapes directly from point clouds using a learned feed-forward model to estimate the shape as a voxel-grid \cite{dai2017shape, han2017high}, point cloud \cite{yuan2018pcn, zhou2022seedformer, chen2023anchorformer}, or neural field \cite{yan2022shapeformer, mittal2022autosdf, li2025noksr, williams2022neural, erler2020points2surf, mescheder2019occupancy, Boulch:2022:POCO}.
Learning-based methods can learn to handle complex measurement noise but typically struggle with low surface detail, often estimating overly smooth shapes \cite{harenstam2024diffcd}.

\inparagraph{Learned prior.}
Some methods combine the two approaches by separately learning a generic shape prior, which is then used in combination with an optimization-based method at inference time.
A core approach in this category is DeepSDF \cite{park2019deepsdf}, which learns a latent space that can be decoded to SDF values using an MLP.
With KL-regularization, it is then possible to perform maximum a posteriori (MAP) inference over shapes.
Another approach is to use Neural Kernel Fields \cite{williams2022neural, Huang:2023:NKS}, which learn to extract kernel parameters from the point cloud. These kernel features are then converted to SDF values by solving a kernel regression problem.
Our method also fits into the learned prior category, using a 3D latent diffusion model as the learned prior and GG-Langevin for inference.

\subsection{Generative models for 3D shapes}
The advent of large open datasets of 3D shapes \cite{deitke2023objaverse, deitke2023objaversexxl} has enabled scalable generative modeling in the 3D domain.
Following similar trends in the vision domain \cite{rombach2022latentDiffusion}, generative models are typically trained in the latent space of an autoencoder.
For 3D shapes, the predominant autoencoder is 3DShape2VecSet~\cite{zhang20233dshape2vecset} (VecSet), which represents each shape as a set of latent vectors that can be decoded into Signed Distance Field (SDF) values or occupancy.
The VecSet approach has enabled the training of large diffusion models~\cite{zhang2024clay, zhao2023michelangelo, yang2024hunyuan3d, hunyuan3d22025tencent, li2025triposg, lai2025flashVDM} with various conditioning modalities, including sketches, point clouds, and multi-view images.
There have been additional improvements to make the underlying autoencoder more expressive~\cite{chen2025dora, zhang2025lagem} and efficient~\cite{lai2025flashVDM, cho2025codvae, zhang2025lagem}, increasing its applicability for downstream tasks. 
Shape diffusion models can also be explicitly trained to sample complete shapes from partial measurements by conditioning the sampling on incomplete point clouds~\cite{chu2023diffcomplete, zhang20233dshape2vecset}.
However, such approaches lack strong measurement consistency and require task-specific training.

\inparagraph{Diffusion guidance.}
Several works use the gradients of a loss function to guide the sampling trajectories of a diffusion model at inference time.
This approach was first used in the image domain as classifier guidance \cite{dhariwal2021diffusion}, in which case the loss function is cross-entropy over the classifier logits.
However, the practical use of classifier guidance is limited because it requires a loss function that is valid for noisy data.
To circumvent this issue, DPS \cite{chung2022dps} and LGD \cite{song2023loss} propose denoising the sample with Tweedie's formula at each step of the sampling trajectory and computing the loss gradient on the denoised sample.
DAPS \cite{zhang2025improving} improves on the quality of guided samples using annealed Langevin sampling.
DPS has been applied to point cloud reconstruction \cite{mobius2024diffusion}, but, as we demonstrate in our experiments, it does not extend well to latent shape reconstruction.

\section{Generative shape reconstruction with GG-Langevin}
\label{sec:method}
The task of shape reconstruction is to estimate a complete shape $\shape$ from a sparse and noisy point cloud measurement $\pointcloud = \{\point_i\}_{i=1}^N$.
In particular, we consider the case where extreme sparsity and incomplete coverage necessitate a data-informed prior to recover the full shape.
We assume access to a diffusion model for sampling from the generic data distribution $p(\latent)$, but that the measurement posterior $p(\latent\sep\pointcloud)$ is unknown. That is, the diffusion model does not take the measurements $\pointcloud$ into account.
We tackle this problem, which we refer to as \textit{generative shape reconstruction}, with a probabilistic approach based on Langevin dynamics \cite{welling2011bayesian, roberts1996exponential, song2019langevin}.
Specifically, we use guided Langevin dynamics, which we find to be an ideal framework for making full use of the diffusion prior $p(\latent)$ while keeping the flexibility of optimization-based methods.
We detail our probabilistic approach and sampling method in \cref{sec:geometric-guidance,sec:hdnd}, and then apply it to the surface reconstruction problem in \cref{sec:shape-reconstruction}.

\subsection{Geometric guidance}
\label{sec:geometric-guidance}
\begin{figure}[t]
\centering 
\includegraphics[width=0.9\textwidth,trim=0 48 0 12,clip]{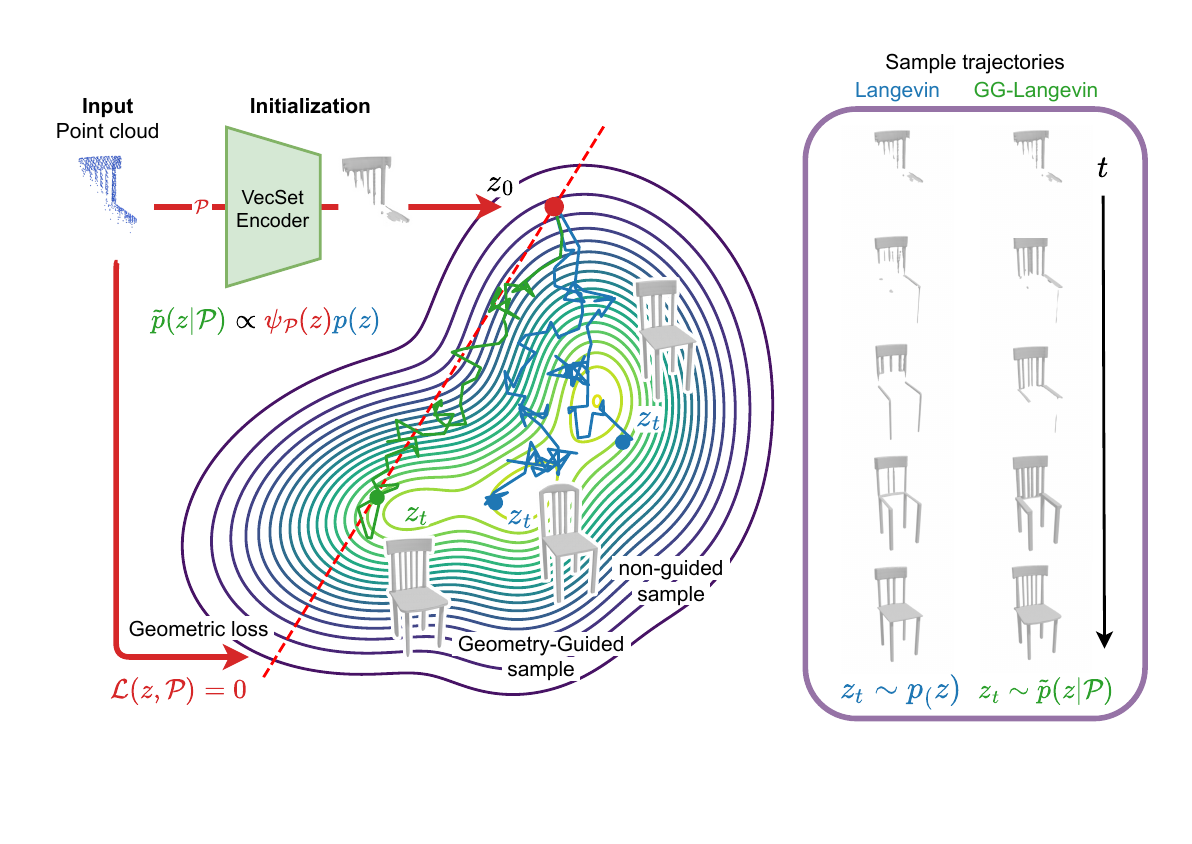}
\vspace{-0.4cm}
\caption{%
    \textcolor{mplblue}{\textbf{Blue trajectories:}} Non-guided Langevin dynamics on the prior distribution $p(\latent)$, initialized at an incomplete shape using the VAE encoder $\latent_0 = \enc(\pointcloud)$. It generates plausible, complete shapes but quickly drifts from the measurements.
    \textcolor{mplgreen}{\textbf{Green trajectory:}} GG-Langevin generatively reconstructs the shape from the input point cloud. By incorporating gradients from a geometric loss $\mathcal{L}(\latent,\pointcloud)$, it keeps the sampling trajectory close to the manifold of measurement-consistent shapes where $\mathcal{L}(\latent,\pointcloud) = 0$ (indicated by the dashed red line). 
    \textbf{On the right:} A side-by-side comparison of sampling trajectories from Langevin dynamics and Geometry-Guided Langevin dynamics.
}
\vspace{-0.5cm}
\label{fig:teaser2}
\end{figure}

A natural way to add measurement consistency to the diffusion prior $p(\latent)$ is to define a geometry-guided shape distribution, which constrains the diffusion prior to shapes that are measurement-consistent:
\eq{
    \label{eqn:gg-density}
    \pguided(\latent\sep\pointcloud) = \frac{1}{\mathcal{Z}(\mathcal{P})}\psi_{\pointcloud}(\latent)\, p(\latent)\,.
}
Here, $\mathcal{Z}(\mathcal{P})$ is a normalization constant and $\psi_{\pointcloud}(\latent) = \exp{(-\invtemp\geomloss{\latent}{\pointcloud})}$ is a per-sample weighting factor based on a geometric loss function $\geomloss{\latent}{\pointcloud}$.
The hyperparameter $\invtemp > 0$ determines how quickly the weighting factor decays to zero as the geometric loss increases.
Intuitively, shapes sampled from $\pguided(\latent\sep\pointcloud)$ satisfy two conditions: they are probable \wrt the prior $p(\latent)$, and they are consistent with the measurements by minimizing $\geomloss{\latent}{\pointcloud}$.
The challenge now is to design an efficient method for sampling from $\pguided(\latent\sep\pointcloud)$.

Existing methods for sampling from $\pguided(\latent\sep\pointcloud)$ are based on solving a reverse-time stochastic differential equation \cite{chung2022dps, song2023loss, zhang2024clay}, starting with random noise $\latent_T\sim\mathcal{N}(0, 1)$ and ending at the guided distribution $\latent_0\sim\pguided(\latent\sep\pointcloud)$.
Intermediate samples are then distributed according to noise-perturbed versions of the guided distribution $z_t\sim \pguided_t(\latent_t|\pointcloud) = \int p_t(\latent_t|\latent)\pguided(\latent\sep\pointcloud)\diff{\latent}$.
However, as the score functions of $\pguided_t$ are unavailable, significant approximations must be applied to obtain a tractable sampling procedure, which ultimately degrades sample quality.
From an algorithmic perspective, starting the sampling process from random noise is also highly impractical, as the loss function $\mathcal{L}(\latent, \pointcloud)$ is only defined for noise-free latents $\latent$.

\subsection{Sampling from $\pguided(\latent\sep\pointcloud)$ with HDND}
\label{sec:hdnd}
As a more appealing alternative, we propose sampling from $\pguided(\latent\sep\pointcloud)$ using a modified version of Langevin dynamics, adapted to take into account the noisy-data score function $s_\sigma(\latent)$, while keeping the benefits of guidance.
Before describing our method, we first briefly review regular (discretized) Langevin dynamics \cite{welling2011bayesian, roberts1996exponential, song2019langevin}. Provided the true (noise-free) score function $s(\latent) = \nabla_\latent\log p(\latent)$ of $p(\latent)$, and a starting point $\latent_0$ with $p(\latent_0) > 0$, samples from $p(\latent)$ can be obtained by iterating the following update rule:
\eq{
\label{eqn:langevin}
        \tilde \latent_t  = \latent_t + \sigma n,
\quad   \latent_{t + 1}   = \tilde \latent_t + \frac{\sigma^2}{2} s(\latent_t),
}
where $n\sim\mathcal{N}(0, 1)$ and $\sigma$ is the noise level.
That is, at every step, the sample is perturbed by noise as well as moved towards the direction of increasing probability.
If the true score function $s(\latent)$ was available, we could therefore sample from $\pguided(\latent\sep\pointcloud)$ by simply plugging $\pguided(\latent\sep\pointcloud)$ into \cref{eqn:langevin}:

\eq{
\label{eqn:gg-langevin-full}
    \tilde \latent_t  = \latent_t + \sigma n,
\quad  \latent_{t+1}     = \tilde \latent_t + \frac{\sigma^2}{2}s(\latent_t) - \guidancestrength\nabla_z\geomloss{\latent_t}{\pointcloud},
}
where $\guidancestrength = \invtemp\frac{\sigma^2}{2}$ is the effective guidance strength.
The key advantage of \cref{eqn:gg-langevin-full} for our purpose is that $\latent_t$ is approximately distributed according to the geometry-guided distribution $\pguided(\latent\sep\pointcloud)$ at every step, provided that the initial sample is sufficiently likely according to the prior distribution.
Critically, this ensures that the gradients $\nabla_z\geomloss{\latent_t}{\pointcloud}$ are always geometrically meaningful without applying any additional denoising to the intermediate states $\latent_t$, a fundamental difference from existing methods for diffusion guidance \cite{zhang2025improving, chung2022dps, song2023loss}.
This makes \cref{eqn:gg-langevin-full} an ideal basis for gradient-based guidance.
However, we still need to account for the fact that only the noisy-data score functions $s_\sigma(\latent)$ are available in practice.

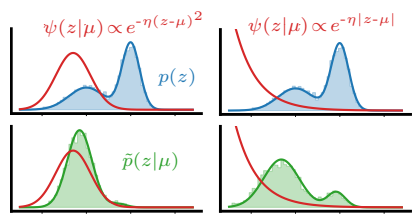
\begin{wrapfigure}{r}{0.45\textwidth}
    \vspace{-2pt}
    \centering
    \begin{tikzpicture}
        \node[inner sep=0] (img) {
            \includegraphics[
                width=\linewidth,
                trim={29mm 0 5mm 24mm},
                clip
            ]{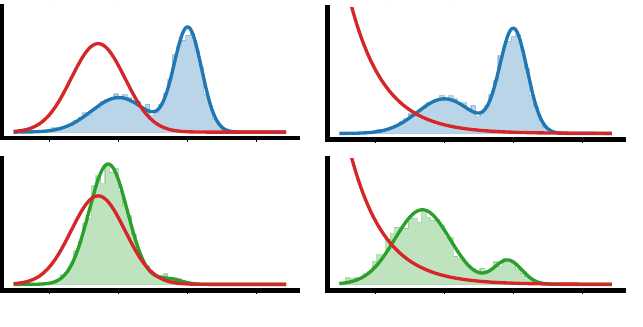}
        };
        \node at ([xshift=-35pt,yshift=0pt] img.north) {\scriptsize\color{mplred} $\psi(\latent\sep\mu)\!\propto\! e^{\text{-}\eta(\latent \text{-} \mu)^2}$};
        \node at ([xshift=40pt,yshift=0pt] img.north) {\scriptsize\color{mplred} $\psi(\latent\sep\mu)\!\propto\! e^{\text{-}\eta|\latent \text{-} \mu|}$};
        \node at ([xshift=-15pt,yshift=20pt] img.center) {\scriptsize\color{mplblue} $p(\latent)$};
        \node at ([xshift=-25pt,yshift=-12pt] img.center) {\scriptsize\color{mplgreen} $\pguided(\latent\sep\mu)$};
    \end{tikzpicture}
    \vspace{-0.9cm}
    \caption{\textbf{Toy example.} Demonstration that our method generates samples from the geometry-guided distribution $\pguided(\latent\sep\pointcloud)$. \textcolor{mplblue}{\textbf{Blue:}} Data distribution. \textcolor{mplred}{\textbf{Red:}} Two variants of guidance weight. \textcolor{mplgreen}{\textbf{Green:}} Geometry-guided product distribution. Solid lines indicate the predicted closed-form distributions. The histograms show samples generated using regular Langevin dynamics and GG-Langevin, respectively. We train an MLP diffusion model on samples from a bimodal Gaussian $p(z)$. The samples closely follow the predicted distributions.}
    \label{fig:histograms}
    \vspace{-0.5cm}
\end{wrapfigure}

We take into account the noisy-data score function by using a recently developed half-denoising variant of Langevin dynamics \cite{hyvarinen2024noise}.
Namely, it turns out that samples from $p(\latent)$ can be obtained using the noisy-data score function as well by simply replacing $s(\latent_t)$ with $s_\sigma(\tilde \latent_t)$ in \cref{eqn:langevin}, where $\sigma$ is a sufficiently small fixed noise level.
The name ``half-denoising'' reflects the fact that the resulting update rule corresponds to subtracting half of the noise estimated from the noised latent $\tilde\latent_t$ at every sampling step.
Unfortunately, applying half-denoising directly to \cref{eqn:gg-langevin-full} is not practical either, as it requires computing the score function of the noise-perturbed distribution $\pguided_t(\latent\sep\pointcloud)$.
So, to arrive at a practical sampling method, we propose using a hybrid ``Half-Denoising-No-Denoising'' (HDND) Langevin update rule, where we apply the half-denoising update rule only to the data term while keeping the guidance term unchanged.
This way, the diffusion model always operates on noised latents $\tilde\latent_t$ (half-denoising), while the geometric loss always operates on denoised latents $\latent_t$ (no-denoising):
\eq{
\label{eqn:gg-langevin-half}
        \tilde \latent_t  = \latent_t + \sigma n,
\quad   \latent_{t+1}     = \tilde \latent_t + \frac{\sigma^2}{2}s_\sigma(\tilde\latent_t) - \guidancestrength\nabla_{\latent}\geomloss{\latent_t}{\pointcloud}.
}
In contrast to \cref{eqn:gg-langevin-full}, this update rule can be implemented in practice by estimating $s_\sigma(z)$ with a diffusion model.
As $\geomloss{\latent}{\pointcloud}$ in our case is a geometric loss, we refer to \cref{eqn:gg-langevin-half} as Geometry-Guided Langevin dynamics (GG-Langevin). 

GG-Langevin is effectively the sum of two separate Langevin processes: one half-denoised Langevin process on  $p(\latent)$, ensuring consistency with the data distribution, and one regular Langevin process on $\psi_{\pointcloud}(\latent)$ for geometric guidance, ensuring measurement consistency.
While this hybrid approach only approximately samples from the guided distribution $\pguided(\latent\sep\pointcloud)$, we find that the approximation holds remarkably well in practice.
We demonstrate this with a 1D toy example in \cref{fig:histograms} for the loss functions $\geomloss{z}{\mu} = (z - \mu)^2$ and $\geomloss{z}{\mu}=|z - \mu|$ with a small MLP diffusion model trained to fit a bimodal Gaussian distribution. Note that the samples generated with GG-Langevin (green histograms) closely match the predicted product distribution (solid green lines) in both cases.

As an inference method, GG-Langevin has several key advantages: it samples from a well-defined distribution, does not require noise-level scheduling by keeping $\sigma$ constant, and can be efficiently initialized from an initial estimate $z_0$ (\eg, via the point cloud encoder, with $\latent_0 = \enc(\pointcloud)$).
In practice, the close resemblance of \cref{eqn:gg-langevin-half} to gradient-based optimization also allows us to leverage modern neural network optimizers for efficient convergence.
With this in mind, we adapt GG-Langevin for surface reconstruction in the following section.

\begin{algorithm}[t]
    \caption{Geometry-Guided Langevin dynamics}
    \label{alg:shape_completion}
    \begin{algorithmic}
        \Require Point cloud $\pointcloud$, autoencoder ($\enc, \dec$), score model $s_{\sigma}$, guidance strength $\beta$, (optionally scheduled) noise 
        $\sigma_t$.
        \State Initialize $z_0 = \enc(\pointcloud)$
        \State Initialize Adam optimizer $\mathcal{O}_{\text{adam}}$ with learning rate $\beta$
        \For{$i = 0, \ldots, N - 1$}
            \State Add noise: $\tilde \latent_t  = \latent_t + \sigma_t n$
            \State Half-denoising: $\hat z_t = \tilde \latent_t + \frac{\sigma_t^2}{2}s_{\sigma_t}(\tilde\latent_t)$
            \State Geometric guidance (\ref{eqn:geom_loss}, \ref{eqn:geom_loss_per_component}): $g_t =\nabla_{\latent}\geomloss{\latent_t} {\pointcloud}$
            \State Adam update rule $\latent_{t+1} = \mathcal{O}_{\text{adam}} (\hat z_t, g_t)$
        \EndFor
    \end{algorithmic}
    \Return $\latent_{N}$
\end{algorithm}

\begin{figure}[t]
\centering
\begin{tikzpicture}
\node[inner sep=0] (img) {\includegraphics[width=\linewidth]{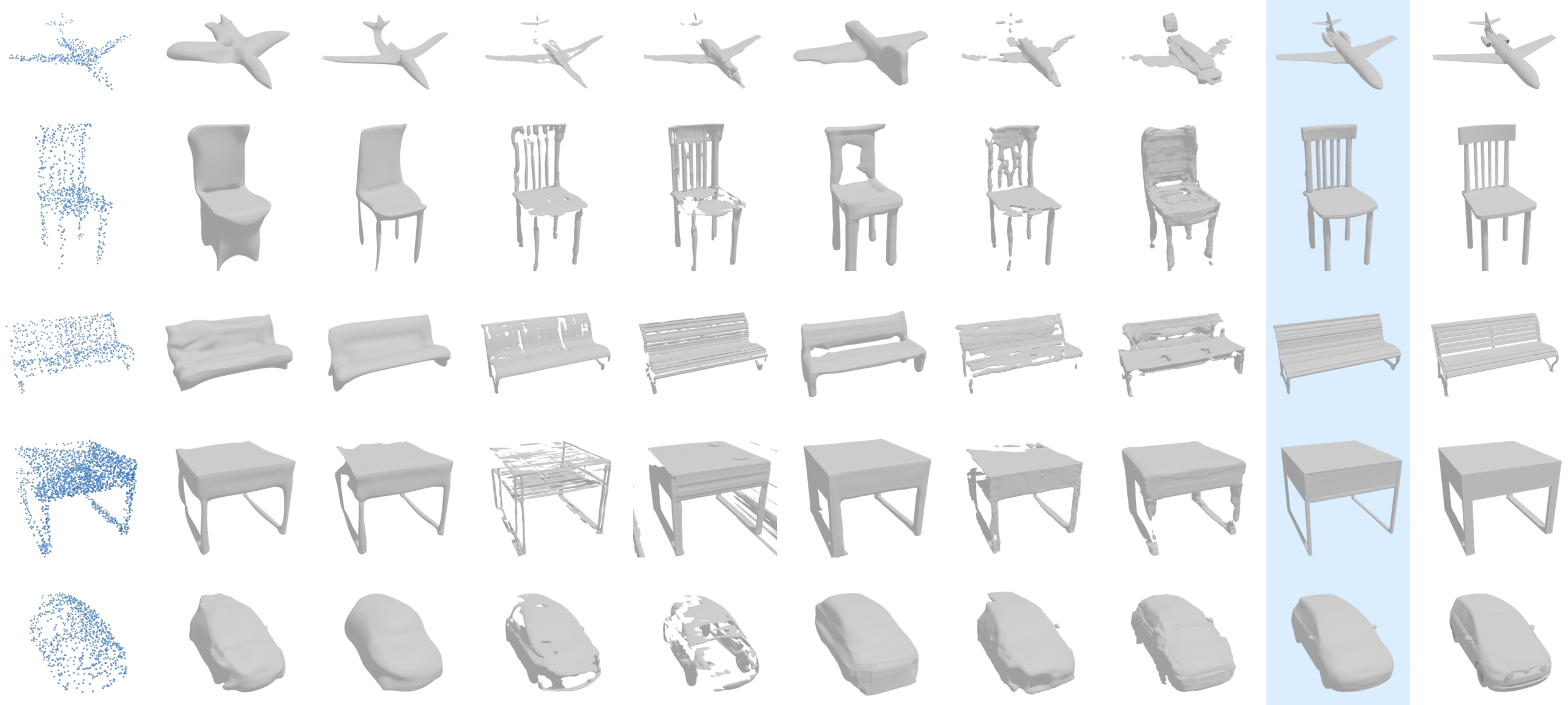}};
\foreach \x/\lab in {
    0.045872/Input,
    0.146789/IGR,
    0.247706/DiffCD,
    0.348624/3DILG,
    0.449541/VecSet,
    0.550459/ShapeFormer,
    0.651376/NKSR,
    0.752294/DeepSDF,
    0.853211/\textbf{Ours}, 
    0.954128/GT
} {
    \path (img.north west) -- (img.north east) coordinate[pos=\x] (t);
    \node[anchor=south, text height=1.2ex, text depth=0.25ex] at (t) {\fontsize{4.0pt}{4.0pt}\selectfont \lab};
}
\end{tikzpicture}
\caption{\textbf{Reconstruction results on sparse point clouds.} Provided sparse point cloud scans as input, GG-Langevin recovers the complete surface and fine structures, significantly improving the reconstruction accuracy in comparison to previous work.}
\label{fig:shapenet-results-sparse}
\end{figure}
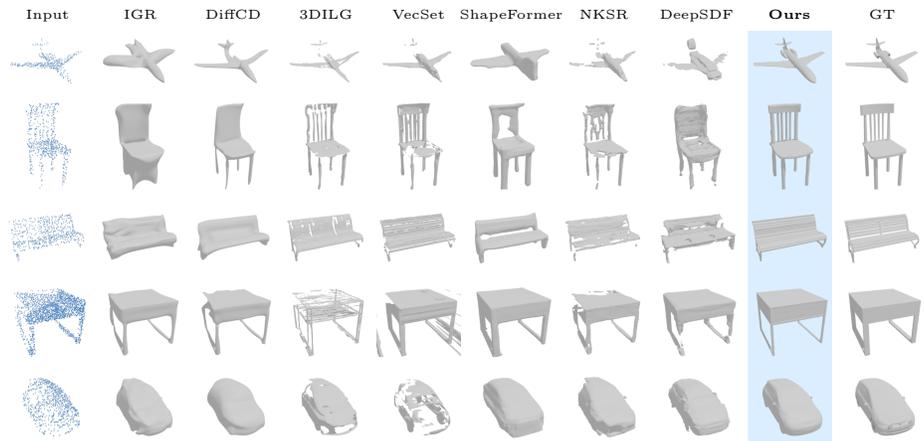

\subsection{Shape reconstruction}
\label{sec:shape-reconstruction}
To apply GG-Langevin for surface reconstruction, we first need to define the shape parametrization $z$ and the geometric loss function $\mathcal{L}(\latent,\pointcloud)$ for guidance.
As shape parametrization, we use the latent space of a VecSet variational autoencoder (VAE) \cite{zhang20233dshape2vecset}, which is a commonly used parametrization for large-scale 3D generative models \cite{zhang2024clay, zhao2023michelangelo, yang2024hunyuan3d, hunyuan3d22025tencent, li2025triposg, lai2025flashVDM}.
An encoder $\enc$ is trained to map complete point clouds $\pointcloud$ to latent vectors: $z = E(\pointcloud)$, and a decoder $\dec$ is trained to predict the corresponding SDF values $\dec(\latent,\point)$ at query points $\point\in\mathbb{R}^3$.
Via the decoder, each latent vector therefore represents a surface, which is obtained by  extracting the 0-level set $S_\latent = \{\point: \dec(\latent,\point) = 0\}$.
As the decoder is differentiable with respect to both arguments, we can apply well-established loss functions from neural implicit surface reconstruction to optimize for the latent $\latent$ that best matches the input point cloud.
In particular, we use the IGR loss function \cite{gropp2020implicit}:
\eq{
\label{eqn:geom_loss}
    \geomloss{\latent}{\pointcloud} =
    \surfloss(\latent, \pointcloud) +
    \lambda \eikloss(\latent),
}
where
\eq{
\label{eqn:geom_loss_per_component}
    \surfloss(\latent, \pointcloud)
    &= \frac{1}{N}\sum_{i=1}^N|\dec(\latent, \point_i)|,
\\    \eikloss(\latent)
    &= \mathbb{E}_{\point\sim p_\text{eikonal}}\big(\|\nabla_\point \dec(\latent, \point)\| - 1\big)^2.
}
Here, $\surfloss$ ensures that the shape approximately fits the point cloud points $\point_i$, and $\eikloss$ pushes the SDF to satisfy the eikonal equation everywhere in the bounding volume $\Omega = [-1,1]^3$ \cite{gropp2020implicit}.
Although various extensions to \cref{eqn:geom_loss} have been proposed  \cite{sitzmann2019siren,harenstam2024diffcd,BenShabat2021DiGS,wang2023aligningHessianGradient,zixiong23neuralsingular,wang2025hotspot}, we find that the diffusion prior on its own provides sufficient regularization for highly accurate surface reconstruction.

We provide a visual overview of our method in \cref{fig:teaser2}.
The sampling trajectory starts at the initial estimate $z_0$, which is typically inaccurate and therefore lies in a low-probability region.
Then, at each iteration, the score function term completes the shape by pulling the sample towards the shape distribution, while the guidance term maintains measurement consistency, \ie, $\geomloss{\latent}{\pointcloud}\approx0$.

\subsection{Implementation details}
\label{sec: implementation details main paper}
\inparagraph{Encoder initialization.} Although the encoder $\enc$ is trained on complete and noise-free point clouds, we find that it often provides a reasonable estimate even when the point cloud is noisy and incomplete. It can therefore naturally be used to initialize our method, reducing the total number of iterations required to converge to the complete shape. That is, we initialize with $\latent_0 = \enc(\pointcloud)$.

\inparagraph{Guidance strength.} The guidance strength $\guidancestrength$ is a tunable parameter, determining the relative strength of the geometric prior.
We note that \cref{eqn:gg-langevin-half} resembles gradient descent on the effective loss function
$-\log p(\latent) + \guidancestrength\geomloss{\latent}{\pointcloud}$
with an added noise term.
Consequently, we use the Adam optimizer \cite{kingma2014adam} to compute gradient updates.
See \cref{alg:shape_completion} for a summary of our approach.

\begin{figure}[t]
    \centering 
    \includegraphics[width=\textwidth,trim={10mm 0 5mm 0},clip]{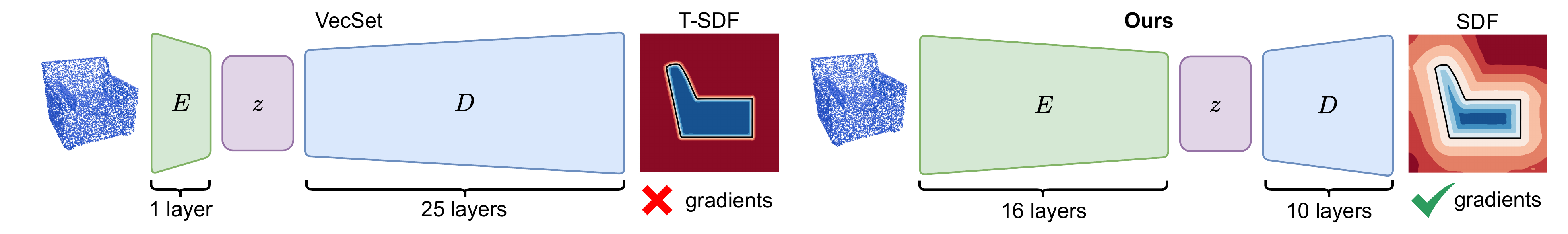}
    \caption{\textbf{Design choices for the autoencoder.} We reduce the size of the decoder while maintaining the overall size of the autoencoder. Further, we predict a full SDF instead of a Truncated SDF (T-SDF) to enable gradients for the geometric prior.}
    \label{fig:enc_dec_changes}
\end{figure}

\inparagraph{Efficient autoencoder design.}
\label{sec:efficient-latents}
At each step of GG-Langevin, we need to compute gradients of the decoder $\dec(\latent,\point)$.
This imposes some new requirements on the decoder design, namely that it should be differentiable and as efficient as possible.
In contrast, the encoder $\enc(\pointcloud)$ is only used once at initialization.
This raises some issues with existing VecSet autoencoder designs~\cite{li2025triposg, hunyuan3d22025tencent}, as they employ small encoders consisting of a single cross-attention layer and, correspondingly, large decoders.
The large decoder naturally makes the gradients $\nabla_\latent\geomloss{\latent}{\pointcloud}$ computationally expensive.
We alleviate these issues by moving the encoder-decoder bottleneck to a later layer.
This leads to a more expressive latent space due to the larger encoder, while also significantly reducing the time required for propagating gradients from the geometric loss to the latent space.
Second, we train the autoencoder to predict full, untruncated SDF values, ensuring that the eikonal constraint is satisfied everywhere in $\Omega$.
We provide a high-level overview of the autoencoder architecture in \cref{fig:enc_dec_changes} and evaluate our performance for different bottleneck positions in \cref{sec:autoencoder-ablation}.

\begin{table}[t]
    \footnotesize
    \centering
    \caption{\textbf{Shape reconstruction.} Comparison of different methods on ShapeNet categories for sparse and incomplete scans. Lower is better for both Chamfer Distance $\times 10^{2}$ (CD) and Chamfer Angle (CA) in degrees. We highlight the best-performing method in each object category in bold and the second-best underlined.
    }
    \setlength{\tabcolsep}{4pt}
    \renewcommand{\arraystretch}{1.1}
    \resizebox{\textwidth}{!}{%
    \begin{tabular}{llcccccccccccccccc}
        & &
        \multicolumn{8}{c}{\textbf{Sparse Scans}} &
        \multicolumn{8}{c}{\textbf{Incomplete Scans}} \\
        \cmidrule(lr){3-10} \cmidrule(lr){11-18}
        & 
        & \multicolumn{2}{c}{Cars}
        & \multicolumn{2}{c}{Airplanes}
        & \multicolumn{2}{c}{Tables}
        & \multicolumn{2}{c}{Chairs}
        & \multicolumn{2}{c}{Cars}
        & \multicolumn{2}{c}{Airplanes}
        & \multicolumn{2}{c}{Tables}
        & \multicolumn{2}{c}{Chairs} \\
        \cmidrule(lr){3-4} \cmidrule(lr){5-6} \cmidrule(lr){7-8} \cmidrule(lr){9-10}
        \cmidrule(lr){11-12} \cmidrule(lr){13-14} \cmidrule(lr){15-16} \cmidrule(lr){17-18}
        & \textbf{} &
        CD & CA 
        & CD & CA 
        & CD & CA 
        & CD & \multicolumn{1}{c}{CA} 
        & CD & CA 
        & CD & CA 
        & CD & CA 
        & CD & CA 
        \\
        \midrule

        \multirow{2}{*}{Optimization} & IGR \cite{gropp2020implicit}
	& \underline{1.07} & \underline{27.7} 
	& 2.80 & 33.4 
	& 2.36 & 25.1 
	& 2.52 & 29.1 
	& 4.47 & 32.8 
	& 3.82 & 31.7 
	& 6.04 & 29.2 
	& 5.38 & 30.4 
	\\
	& DiffCD \cite{harenstam2024diffcd}
	& 1.22 & 29.3 
	& \underline{0.88} & \underline{25.7} 
	& 1.48 & 21.0 
	& 1.58 & 24.1 
	& 5.40 & 33.1 
	& 3.26 & 28.4 
	& 6.02 & 23.5 
	& 5.13 & 26.0 
	\\
    \midrule
	\multirow{3}{*}{Learning-based} & 3DILG \cite{zhang20223dilg}
	& 1.68 & 30.6 
	& 1.01 & 25.9 
	& 1.37 & 23.1 
	& 1.38 & 26.7 
	& 5.50 & 32.4 
	& 4.40 & 24.3 
	& 4.80 & 18.2 
	& 4.51 & \underline{22.8} 
	\\
	& VecSet \cite{zhang20233dshape2vecset}
	& 1.81 & 32.5 
	& 0.99 & 26.8 
	& \underline{1.29} & \underline{17.7} 
	& \underline{1.23} & 22.6 
	& 5.55 & 36.1 
	& 3.36 & \underline{23.1} 
	& 5.31 & 18.9 
	& 4.96 & 23.8 
	\\
	& ShapeFormer \cite{yan2022shapeformer}
	& 1.92 & 32.0 
	& 2.18 & 33.5 
	& 2.10 & 20.2 
	& 2.66 & 28.6 
	& \underline{2.37} & \underline{30.6} 
	& \underline{2.75} & 34.0 
	& 2.77 & \underline{20.4} 
	& 4.11 & 32.8 
	\\
    \midrule
    \multirow{3}{*}{Learned prior} & DeepSDF \cite{park2019deepsdf}
	& 1.26 & 34.4 
	& 1.55 & 39.5 
	& 1.51 & 25.8 
	& 1.65 & 30.4 
	& 3.83 & 37.9 
	& 3.41 & 42.0 
	& \underline{2.69} & 25.4 
	& \underline{2.44} & 30.9 
	\\
	& NKSR \cite{Huang:2023:NKS}
	& 1.17 & 28.6 
	& 1.26 & 29.7 
	& 1.44 & 18.6 
	& 1.31 & \underline{21.5} 
	& 4.57 & 31.6 
	& 3.62 & 29.1 
	& 5.16 & 21.8 
	& 4.32 & 23.4 
	\\
	& Ours 
	& \textbf{0.88} & \textbf{25.4} 
	& \textbf{0.63} & \textbf{17.7} 
	& \textbf{1.22} & \textbf{14.3} 
	& \textbf{1.04} & \textbf{17.0} 
	& \textbf{0.84} & \textbf{23.3} 
	& \textbf{1.24} & \textbf{17.6} 
	& \textbf{1.61} & \textbf{15.0} 
	& \textbf{1.95} & \textbf{19.2} 
	\\
    \bottomrule
    \end{tabular}%
    }
    \label{tab:shapenet_results}
\end{table}

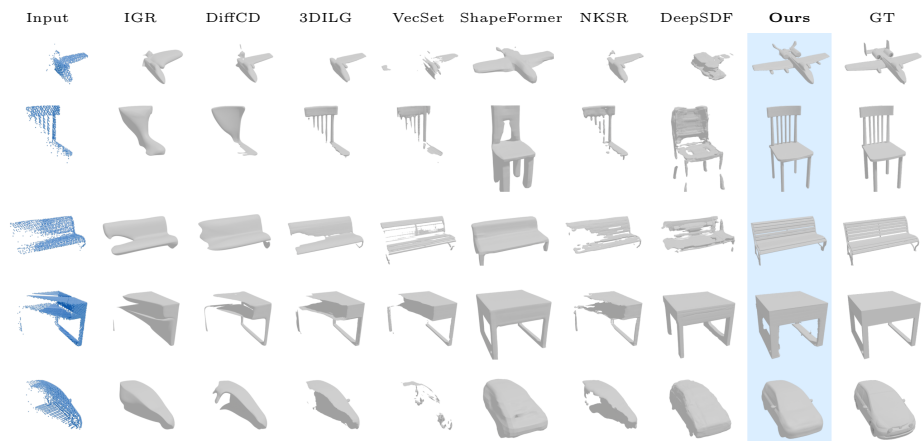
\begin{figure}[t]
\centering
\begin{tikzpicture}
\node[inner sep=0] (img) {\includegraphics[width=\linewidth]{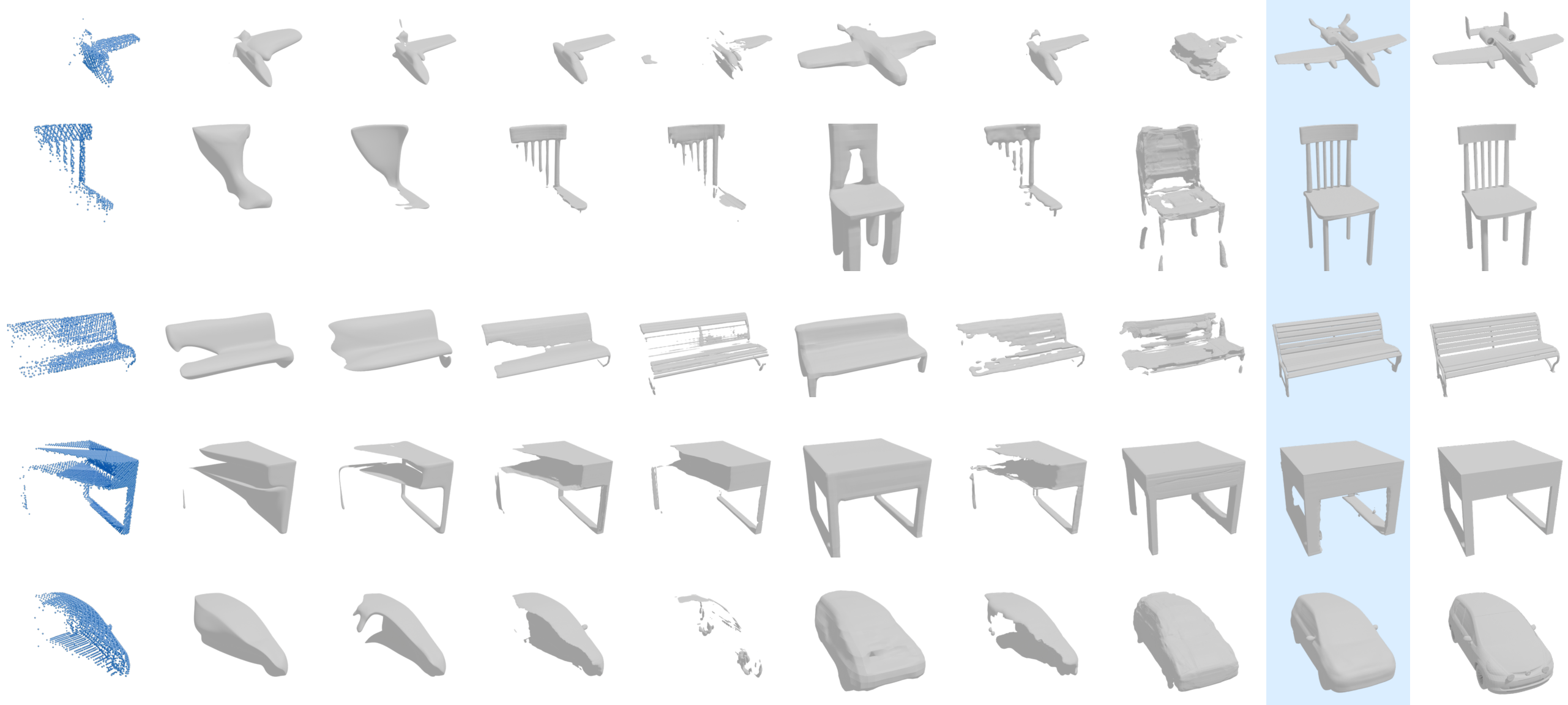}};
\foreach \x/\lab in {
    0.045872/Input,
    0.146789/IGR,
    0.247706/DiffCD,
    0.348624/3DILG,
    0.449541/VecSet,
    0.550459/ShapeFormer,
    0.651376/NKSR,
    0.752294/DeepSDF,
    0.853211/\textbf{Ours}, 
    0.954128/GT
} {
    \path (img.north west) -- (img.north east) coordinate[pos=\x] (t);
    \node[anchor=south, text height=1.2ex, text depth=0.25ex] at (t) {\fontsize{4.0pt}{4.0pt}\selectfont \lab};
}
\end{tikzpicture}
\vspace{-0.5cm}
\caption{\textbf{Reconstruction results on incomplete point clouds.} Despite incomplete point cloud scans as input, GG-Langevin recovers the missing structure with prior-consistent geometry. By comparison, previous work either struggles to complete the geometry or hallucinates implausible completions.}
\label{fig:shapenet-results-partial}
\end{figure}

\section{Experiments}
\label{sec:experiments}
\inparagraph{Shape reconstruction benchmark.} We evaluate surface reconstruction under two settings: \emph{sparse} point cloud scans with noise and \emph{incomplete} point cloud scans with large missing regions.
In both cases, we generate point cloud scans using a realistic scanning procedure similar to Erler \etal \cite{erler2020points2surf}.
We use various types of shapes (Cars, Airplanes, Tables, and Chairs) from ShapeNet~\cite{chang2015shapenet}.
The resulting point clouds exhibit spatially varying sparsity and occasionally contain occluded regions.
To generate incomplete scans, we randomly select a plane normal and an offset, and remove points on one side of the plane with a probability that exponentially decays \wrt the distance to the plane.
This makes the generative prior essential to complete the shape.
We demonstrate in \cref{sec:results} that no existing method for surface reconstruction consistently performs well across this challenging benchmark, a gap that our method closes.

\inparagraph{Autoencoder training.}
We train our rebalanced VAE from scratch with full untruncated SDF outputs on all ShapeNet~\cite{chang2015shapenet} classes. The data and training pipeline is based on VecSet~\cite{zhang20233dshape2vecset}, and we use the L1 loss on the ground-truth SDF values along with an eikonal loss \cite{gropp2020implicit}.
Apart from our rebalanced bottleneck, we also evaluate our method with two other bottleneck positions in \cref{sec:autoencoder-ablation}.

\inparagraph{Diffusion model training.}
To approximate the noisy-data score functions $s_\sigma(\latent)$ of the shape prior $p(\latent)$, we train a diffusion model on the latent space of the rebalanced VAE.
See \cref{sec:additional-implementation-details} for implementation details.
We train our diffusion model without class conditioning, so the object category is unknown at inference time. 
We investigate adding class conditioning in \cref{sec:class-conditioning}.

\inparagraph{Hyperparameters.} As base settings, we use $N=2000$ iterations of GG-Langevin, with $\sigma = 0.05$ (constant), $\beta = 0.03$, and $\lambda = 0.1$.
We investigate the impact of $\sigma$ and $\beta$ on reconstruction quality in \cref{sec:hyperparameter-ablation}.
For incomplete point clouds, we find that the sampling trajectory can get stuck on incomplete shapes if $\sigma$ is chosen too small.
We therefore anneal the noise level from $\sigma_\text{max}=0.2$ to $\sigma_\text{min}=0.02$ over 4000 iterations with a cosine schedule, followed by 1000 iterations with constant $\sigma=0.02$ ($N=5000$ iterations in total).
We evaluate the impact of the annealing schedule in \cref{sec:noise-schedule-ablation}.
We also investigate adding the off-surface loss from Sitzmann \etal \cite{sitzmann2019siren} in \cref{sec:siren-weight-comparison}.

\begin{figure}[t]
\centering
\begin{minipage}{0.52\textwidth}
    \centering
    \begin{tikzpicture}
    \node[inner sep=0] (img) {\includegraphics[width=\linewidth]{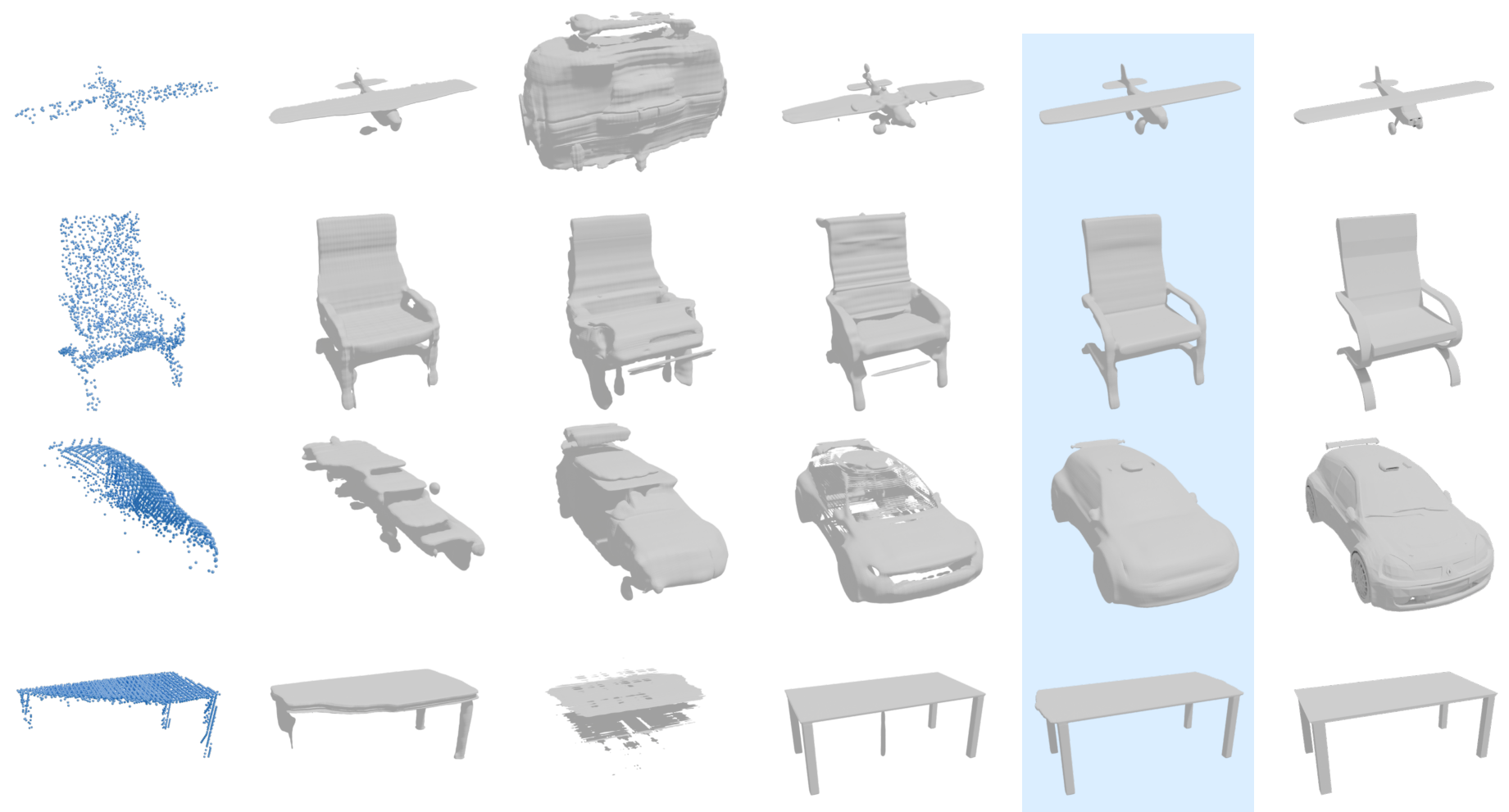}};
    \foreach \x/\lab in {
    0.076923/Input,
    0.246154/{MAP},
    0.415385/{DPS},
    0.584615/{DAPS},
    0.753846/\textbf{Ours},
    0.923077/GT
} {
  \path (img.north west) -- (img.north east) coordinate[pos=\x] (t);
  \node[anchor=south, text height=1.2ex, yshift=-3pt, text depth=0.25ex] at (t) {\fontsize{4.0pt}{4.0pt}\selectfont \lab};
}
\foreach \y/\lab in {
    0.246137/{Sparse},
    0.720613/{Incomplete}
} {
  \path (img.north west) -- (img.south west) coordinate[pos=\y] (l);
  \node[anchor=center, xshift=-3.5pt, rotate=90, text height=1.2ex, text depth=0.25ex] at (l) {\fontsize{4.0pt}{4.0pt}\selectfont \lab};
}
    \end{tikzpicture}
    \label{fig:sampler}
\end{minipage}\hfill
\begin{minipage}{0.42\textwidth}
    \centering
    \footnotesize
    \vspace*{\fill}
    \setlength{\tabcolsep}{3pt}
    \renewcommand{\arraystretch}{1.05}
    \resizebox{\linewidth}{!}{%
    \begin{tabular}{lcccc}
        & \multicolumn{2}{c}{\textbf{Sparse}} & \multicolumn{2}{c}{\textbf{Incomplete}} \\
        \cmidrule(lr){2-3} \cmidrule(lr){4-5}
        \textbf{Sampler} & CD & CA & CD & CA \\
        \midrule
        MAP
	& \underline{1.00} & \underline{21.6} 
	& 3.86 & 37.5 
	\\
	DPS \cite{chung2022dps}
	& 3.26 & 38.7 
	& 4.04 & 37.8 
	\\
	DAPS \cite{zhang2025improving}
	& 1.04 & 23.2 
	& \underline{1.55} & \underline{19.5} 
	\\
        Ours
	& \textbf{0.95} & \textbf{18.6} 
	& \textbf{1.41} & \textbf{18.8} 
	\\
        \bottomrule
    \end{tabular}%
    }
    \label{tab:sampler}
\end{minipage}
\vspace{-0.5cm}
\caption{\textbf{Sampler ablation.} Performance comparison of methods for sampling from the geometry-guided distribution shape
 $\pguided(\latent\sep\pointcloud)$. We use the same latent space and loss function for all methods. The table on the right shows the Chamfer Distance (CD) and Chamfer Angle (CA) averaged across all object categories.}
\label{fig:sampler-ablation}
\end{figure}
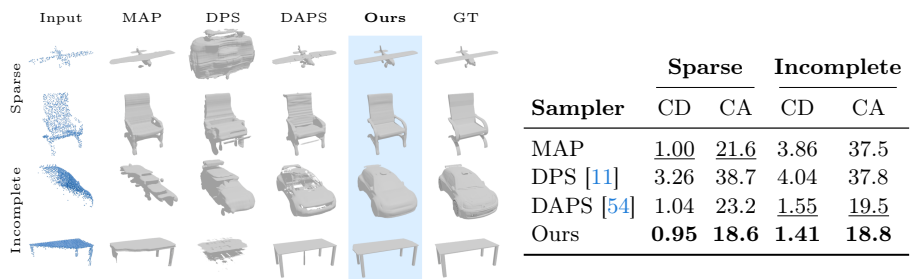

\inparagraph{Baselines.}
We compare our method against the state-of-the-art methods for surface reconstruction.
For optimization-based methods, we compare against IGR \cite{gropp2020implicit}, which optimizes $\geomloss{\latent}{\pointcloud}$ with an MLP parametrization. We also evaluate DiffCD \cite{harenstam2024diffcd}, which extends IGR with a differentiable Chamfer distance to prevent spurious surfaces.
We use the ``medium-noise'' settings from DiffCD \cite{harenstam2024diffcd} for both methods, as we find it provides the best results.
We also compare against learning-based methods ShapeFormer \cite{yan2022shapeformer}, 3DILG \cite{zhang20223dilg}, and the VecSet VAE \cite{zhang20233dshape2vecset} (without diffusion), as well as prior-based methods DeepSDF \cite{park2019deepsdf} (which we retrain from scratch as the original weights are not available) and NKSR \cite{Huang:2023:NKS}.
\cref{sec:results} presents the results.
We also investigate sampling from $\pguided(\latent\sep\pointcloud)$ with alternate guided sampling methods in \cref{sec:sampler-ablation}.

\subsection{Results}
\label{sec:results}
To measure reconstruction quality we compute the Chamfer Distance (CD) and Chamfer Angle (CA) with respect to the ground-truth mesh. 
The Chamfer Angle measures the average angle (in degrees) between the normals of the estimated and ground-truth mesh, using the same point correspondences as for the Chamfer distance.
For SDF-based methods, we decode each shape into a mesh using Marching Cubes \cite{lewiner2003efficientMarchingCubes}.
Our main results are shown in \cref{tab:shapenet_results}. Qualitative results are shown in \cref{fig:shapenet-results-sparse,fig:shapenet-results-partial}.
From \cref{tab:shapenet_results}, we find that our method significantly outperforms all existing methods for shape reconstruction across all categories
\emdash often outperforming the second-best method in each category by a substantial margin.
Furthermore, while some baselines achieve competitive performance on one of the benchmarks (sparse or incomplete), no single baseline is consistently competitive across both benchmarks.
For instance, DiffCD \cite{harenstam2024diffcd}, NKSR \cite{Huang:2023:NKS}, and VecSet \cite{zhang20233dshape2vecset} perform well on sparse scans but fail on incomplete scans.
On the other hand, ShapeFormer \cite{yan2022shapeformer} and DeepSDF \cite{park2019deepsdf} perform well on incomplete scans, but poorly on sparse scans relative to the other methods.
In comparison, our method is uniquely able to make full use of both the measurements (for preserving the original shape) and the prior (for generating missing parts).

\begin{figure}[t]
    \begin{minipage}[c]{0.32\textwidth}
        \centering
        \includegraphics[width=\linewidth]{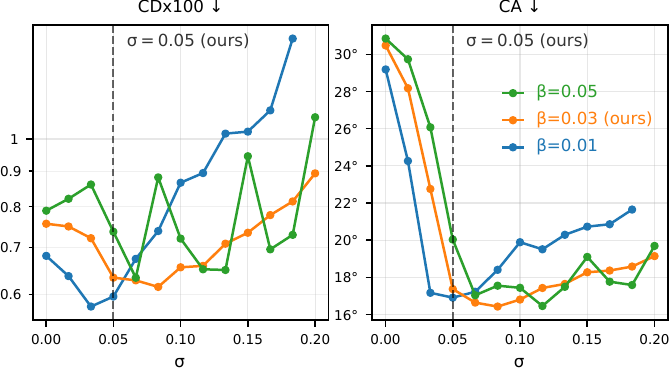}
    \end{minipage}
    \begin{minipage}[c]{0.65\textwidth}
        \centering
        \foreach \lbl in {Input,$\sigma=0$,$\sigma=0.05$,$\sigma=0.2$} {
            \begin{minipage}{0.23\linewidth}\centering \lbl\end{minipage}
        }
        \foreach \img in {
            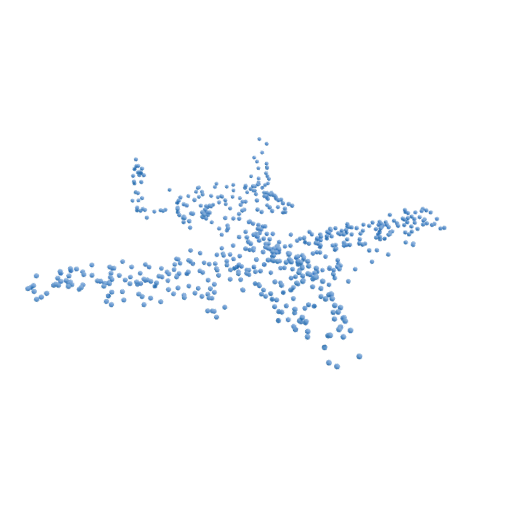,
            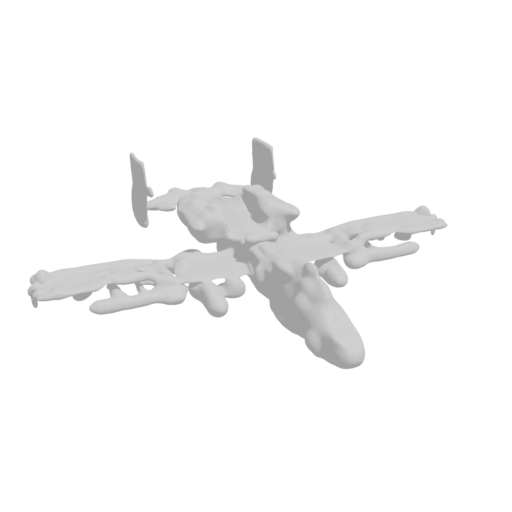,
            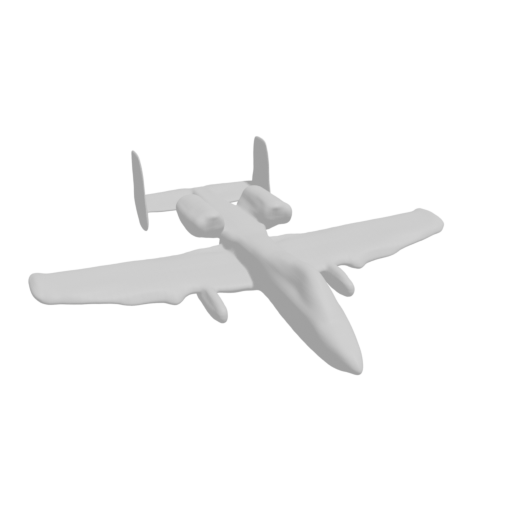,
            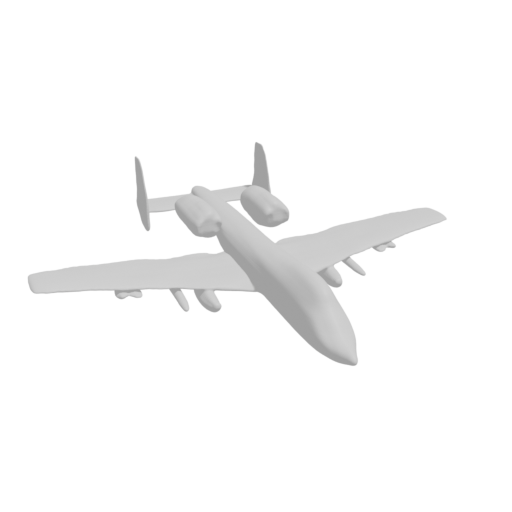
        }{
            \begin{minipage}{0.23\linewidth}
                \centering
                \includegraphics[
                    width=\linewidth,
                    trim={10mm 20mm 20mm 40mm},
                    clip
                ]{\img}
            \end{minipage}
        } 

    \end{minipage}
    \vspace{-0.1cm}
    \caption{
    Impact of noise level $\sigma$ and guidance strength $\beta$ on reconstruction performance.
    If $\sigma$ is too small relative to $\beta$ the shape overfits the noise. Conversely, if $\sigma$ is too large, the shape starts drifting away from the measurements.
    }
    \vspace{-0.3cm}
    \label{fig:ablations}
\end{figure}

\subsection{Sampler ablation}
\label{sec:sampler-ablation}
GG-Langevin is composed of two components: the geometric loss function $\mathcal{L}(\latent,\pointcloud)$ and a method for sampling from the geometry-guided shape distribution $\pguided(\latent\sep\pointcloud)$, defined in \cref{eqn:gg-density}.
As described in \cref{sec:hdnd}, we develop a novel sampling method, HDND, to sample from $\pguided(\latent\sep\pointcloud)$.
To validate HDND, we also investigate sampling from $\pguided(\latent\sep\pointcloud)$ with existing methods DPS \cite{chung2022dps} and DAPS \cite{zhang2025improving}.
For a fair comparison, we adapt both sampling methods to use the encoder initialization as described in \cref{sec:dps-daps-adaptations}.
Since our shapes are parametrized using a VAE, we also investigate using MAP estimation by minimizing $\mathcal{L}_\text{MAP}(\latent,\pointcloud) = \geomloss{\latent}{\pointcloud} + \xi\|z\|^2$ with Adam \cite{kingma2014adam}.

Results for the sampler comparison are shown in \cref{fig:sampler-ablation}.
Our method (GG-Langevin with HDND sampling) consistently outperforms the other sampling methods.
MAP performs surprisingly well in the sparse setting, validating the effectiveness of our VAE design.
However, in the incomplete setting, MAP completely fails to recover a plausible shape in most cases.
We also find that DPS \cite{chung2022dps} struggles to produce reasonable samples.
DPS \cite{chung2022dps} estimates a denoised shape at each step using the Tweedie's formula, but this estimate turns out to be highly inaccurate at early steps (where the noise level is high).
Computing the geometric loss with these inaccurate estimates results in invalid guidance, ultimately causing the sampling trajectory to diverge into blob-like artefacts.
DAPS \cite{zhang2025improving} generally outperforms DPS \cite{chung2022dps}, yet still produces poor surface quality and spurious geometry.
This is because DAPS \cite{zhang2025improving} uses a decoupled approach (alternating denoising and guidance as separate steps), which complicates the tradeoff between measurement consistency and prior consistency.  
In contrast to DPS \cite{chung2022dps} and DAPS \cite{zhang2025improving}, GG-Langevin completely avoids working with high noise levels. It also applies a tightly coupled denoising step and guidance at every iteration, making it significantly easier to maintain the balance between measurement consistency and prior consistency.

\subsection{Autoencoder ablation}
\label{sec:autoencoder-ablation}
We evaluate how our VAE bottleneck position affects reconstruction quality by evaluating alternate VAEs with 25, 10, and 1 decoder layers.
For each setting, we train a VAE on the Chairs category and then a diffusion model on the corresponding latent space.
As shown in \cref{fig:autoencoder-comparison}, reducing the decoder size from 25 to 10 yields a roughly $2\times$ speedup per GG-Langevin iteration.
The reduced decoder size also improves reconstruction results by a significant margin, likely due to the fact that a single encoder layer cannot learn a sufficiently expressive latent space for well-behaved gradients $\nabla_{\latent}\mathcal{L}(\latent,\pointcloud)$.
Reducing the decoder size even further to a single layer yields another $2\times$ speedup, but again comes at the cost of reduced reconstruction performance:
a single decoder layer does not provide sufficient structure for gradient-based guidance.
Overall, our analysis suggests that balancing the number of encoder and decoder layers produces a latent space ideal for both generative modeling and gradient-based guidance.

\subsection{Hyperparameter ablation}
\label{sec:hyperparameter-ablation}
We investigate the impact of the noise level $\sigma$ and the guidance strength $\beta$ in \cref{fig:ablations}.
We evaluate performance on sparse point clouds from the Airplanes category with guidance strengths $\beta = 0.01, 0.03$ and 0.05 for a range of noise levels from $\sigma=0$ to $\sigma=0.2$.
For $\beta=0.03$ (which we use as our base setting), we find that our method generally performs well for noise levels $\sigma$ in the range from $\sigma=0.05$ to $\sigma=0.1$.
The impact of $\sigma$ is particularly evident in the Chamfer Angle metric: using small values of $\sigma$ results in shapes that overfit the point cloud, leading to wobbly surfaces with highly inaccurate normals.
On the other hand, using large values of $\sigma$ results in shapes that drift away from the measurements and overfit the prior.
For the smaller guidance strength $\beta=0.01$, overall performance degrades due to a narrower range of optimal noise levels. For the higher guidance strength $\beta=0.05$, the guidance steps get too large, and performance becomes unreliable (with sporadic jumps in CD and CA).
We use $\sigma=0.05$ for our method, which achieves a good balance between the two extremes.

\begin{figure}[t]
    \centering
    \def\shapeitems{%
        chairs/c54a464d63efcab2c389b3ea958c8248/0mm,%
        chairs/ea77c800bba6566aaf7c7ad2549a1b15/-20mm,%
        chairs/2b6cbad4ba1e9a0645881d7eab1353ba/0mm
    }
    \begin{minipage}[c]{0.63\textwidth}
        \foreach \lineA/\lineB/\labelname/\folder/\prefix [count=\rowcnt] in {%
            {1 Enc. layer}/{25 Dec. layers}/layer0/layer0/,%
            {16 Enc. layers}/{10 Dec. layers}/layer16/layer16/,%
            {25 Enc. layers}/{1 Dec. layer}/layer25/layer25/%
        }{%
            \hbox to \linewidth{%
                \begin{minipage}[c]{0.4\linewidth}
                    \centering
                    \scalebox{1}{\hbox{\lineA}}\\[0.0em]
                    \scalebox{1}{\hbox{\lineB}}
                \end{minipage}
                \begin{minipage}[c]{0.5\linewidth}%
                    \centering%
                    \foreach \cat/\id/\mytrim [count=\i] in \shapeitems {%
                        \includegraphics[width=0.35\linewidth,trim={10mm {\mytrim} 10mm 0},clip]{figures/\cat/\folder/\prefix\id.png}%
                    }%
                \end{minipage}%
            }\par\ifnum\rowcnt<3 \vskip -1em \fi
        }%
    \end{minipage}
    \begin{minipage}[c]{0.36\textwidth}
        \centering
        \footnotesize
        \resizebox{\linewidth}{!}{\begin{tabular}{lccc}
            \toprule
            \textbf{Dec. layers} & \textbf{CD} & \textbf{CA} & \textbf{s/it} \\
            \midrule
            25 (VecSet \cite{zhang20233dshape2vecset}) & 1.28 & 18.7 & 0.21 \\
            10 (Ours) & \textbf{1.12} & \textbf{17.0} & 0.10 \\
            1 & 1.81 & 24.6  & \textbf{0.06} \\
            \bottomrule
        \end{tabular}}\\[0.3em]
    \end{minipage}%
    \hfill
    \vspace{-0.5em}
    \caption{\textbf{Autoencoder ablation.} Reconstruction quality with varying bottleneck position (26 layers total). 
    Using 10 decoder layers balances quality and inference speed.}
    \label{fig:autoencoder-comparison}
\end{figure}

\section{Conclusion}
We present GG-Langevin, a novel shape reconstruction method that integrates the geometric consistency of optimization-based approaches with the powerful priors of large-scale shape diffusion models through a simple yet effective Langevin sampling procedure.
By guiding the sampling trajectories of a diffusion model with the gradients of a geometric loss, GG-Langevin generatively reconstructs shapes that are both plausible and consistent with observed data, without requiring task-specific retraining or direct conditioning.
Extensive experiments under challenging conditions demonstrate that our method achieves state-of-the-art performance in both fidelity and robustness.
Looking forward, our framework pushes the boundaries of generative reconstruction --- solving complex reconstruction problems by combining the strengths of flexible, but generic, generative models with principled measurement consistency.

\bibliographystyle{splncs04}
\bibliography{main}

\clearpage
\setcounter{page}{1}
\appendix
\pagenumbering{roman}

\section{Additional implementation details}
\label{sec:additional-implementation-details}
In this section, we elaborate on the implementation details for training our rebalanced autoencoder and the corresponding diffusion model.

\inparagraph{Autoencoder training.}
We pre-train our rebalanced VAE $(\enc,\dec)$ on the Chairs category for 3400 epochs, followed by 1100 epochs on the full dataset, using a batch size of 256 and a learning rate of 5$\times10^{-5}$.
For each shape in each batch, we sample 1024 points uniformly from the unit cube, 1024 points near the surface of the input mesh, and 2048 points uniformly from the input mesh.
The model is supervised with ground-truth SDF values, eikonal loss, and KL-divergence loss on the bottleneck layer.
We follow the procedure outlined in \cite{zhang20233dshape2vecset} to filter out invalid points on the mesh, such as those in the interior of the shape. We normalize the input points to fit inside the unit sphere.
In total, autoencoder training takes roughly 3 days on 8 A100 GPUs.

\inparagraph{Diffusion model training.}
For the diffusion model, we train a noise predictor $\hat\epsilon_{\sigma,\theta}(\latent)$, where $\theta$ are the model parameters and $\sigma$ is the noise level.
We use the standard noise-prediction objective function
\eq{
    \mathcal{L}_\text{diffusion}(\theta) = \mathbb{E}_{z,\sigma,\epsilon}[w_{\sigma}\|\hat\epsilon_{\sigma,\theta}(\latent + \sigma\epsilon) - \epsilon\|^2],
}
where $z\sim p(z)$, $\sigma\sim\operatorname{LogNormal}(-1.2, 1.2^2)$ (following Zhang~\etal \cite{zhang20233dshape2vecset}), $\epsilon\sim\mathcal{N}(0, I)$, and $w_\sigma$ is a weighting factor dependent on the noise level.
We set $w_\sigma = 1 + \sigma^2$ following Karras~\etal \cite{karras2022elucidating}.
Once the diffusion model has been trained, we can estimate the noisy-data score function $s_\sigma$ using \cite{vincent2011connection}
\eq{
    s_\sigma(\latent) = -\frac{1}{\sigma}\hat\epsilon_{\sigma,\theta}(\latent).
}
Following Karras~\etal \cite{karras2022elucidating}, we parametrize the noise predictor as
\eq{
    \hat\epsilon_{\sigma,\theta}(\latent) = \frac{1}{\sqrt{1 + \sigma^2}}\,\phi_{\sigma,\theta}\!\bigg(\frac{\latent}{\sqrt{1 + \sigma^2}}\bigg) + \frac{\sigma}{1 + \sigma^2}z,
}
where $\phi_{\sigma,\theta}$ is a transformer with 12 layers having 8 heads each of dimension 64.
The final MLP layer is selected such that $\phi_{\sigma,\theta}(\latent) = 0$ at initialization.
Note that the variance of the data satisfies $\sigma_\text{data}^2 = \mathbb{E}[\|z\|^2] \approx 1$ since the autoencoder is trained with KL regularization. 
A simple calculation then reveals that the training objective satisfies $\mathcal{L}_\text{diffusion}(\theta) \approx 1$ at initialization.

We train the diffusion model for 5000 epochs with a batch size of 512 on the Cars, Airplanes, Tables and Chairs categories with a learning rate of $10^{-4}$ and 10\% dropout.
For each shape in each batch, we obtain latents to denoise by sampling 1024 points $\{\point_i\}_{i=1}^{1024}$ from the ground-truth mesh in the same way as for the autoencoder training and obtain $z = \enc(\{\point_i\}_{i=1}^{1024})$ with the encoder.
Training the diffusion model takes roughly 5 days on 8 A100 GPUs.

\section{Failure cases}
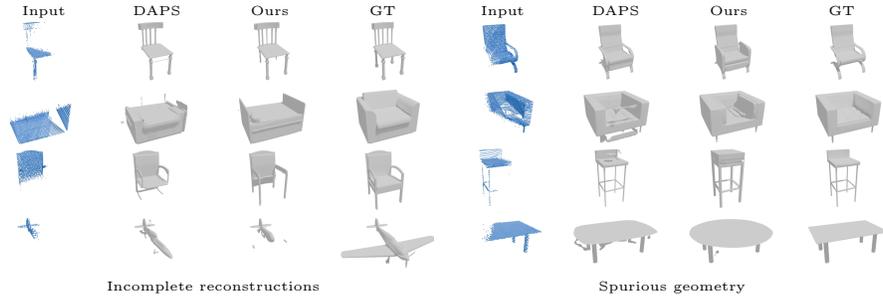
\begin{figure}[t]
\centering
\begin{minipage}{0.48\textwidth}
    \centering
    \begin{tikzpicture}
    \node[inner sep=0] (img) {\includegraphics[width=\linewidth]{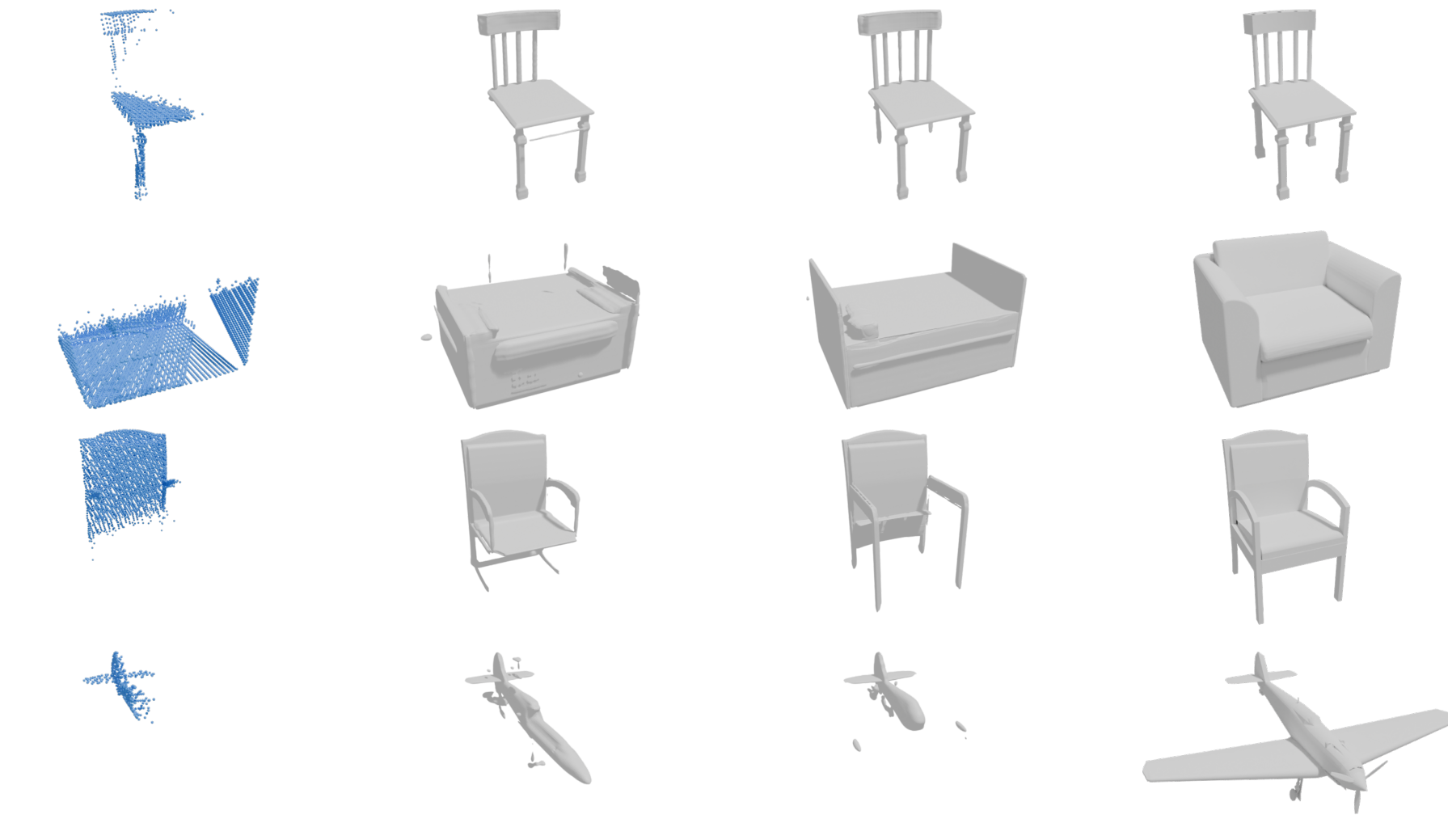}};
\foreach \x/\lab in {
    0.116279/{Input},
    0.372093/{DAPS},
    0.627907/{Ours},
    0.883721/{GT}
} {
  \path (img.north west) -- (img.north east) coordinate[pos=\x] (t);
  \node[anchor=south, text height=1.2ex, yshift=-3pt, text depth=0.25ex] at (t) {\fontsize{4.0pt}{4.0pt}\selectfont \lab};
}
\node[
  anchor=north,
  text height=1.2ex,
  text depth=0.25ex,
  yshift=2pt
] at (img.south) {\fontsize{4.0pt}{4.0pt}\selectfont Incomplete reconstructions};
    \end{tikzpicture}
    \label{fig:failure-c1}
\end{minipage}\hfill
\begin{minipage}{0.48\textwidth}
    \centering
    \begin{tikzpicture}
    \node[inner sep=0] (img) {\includegraphics[width=\linewidth]{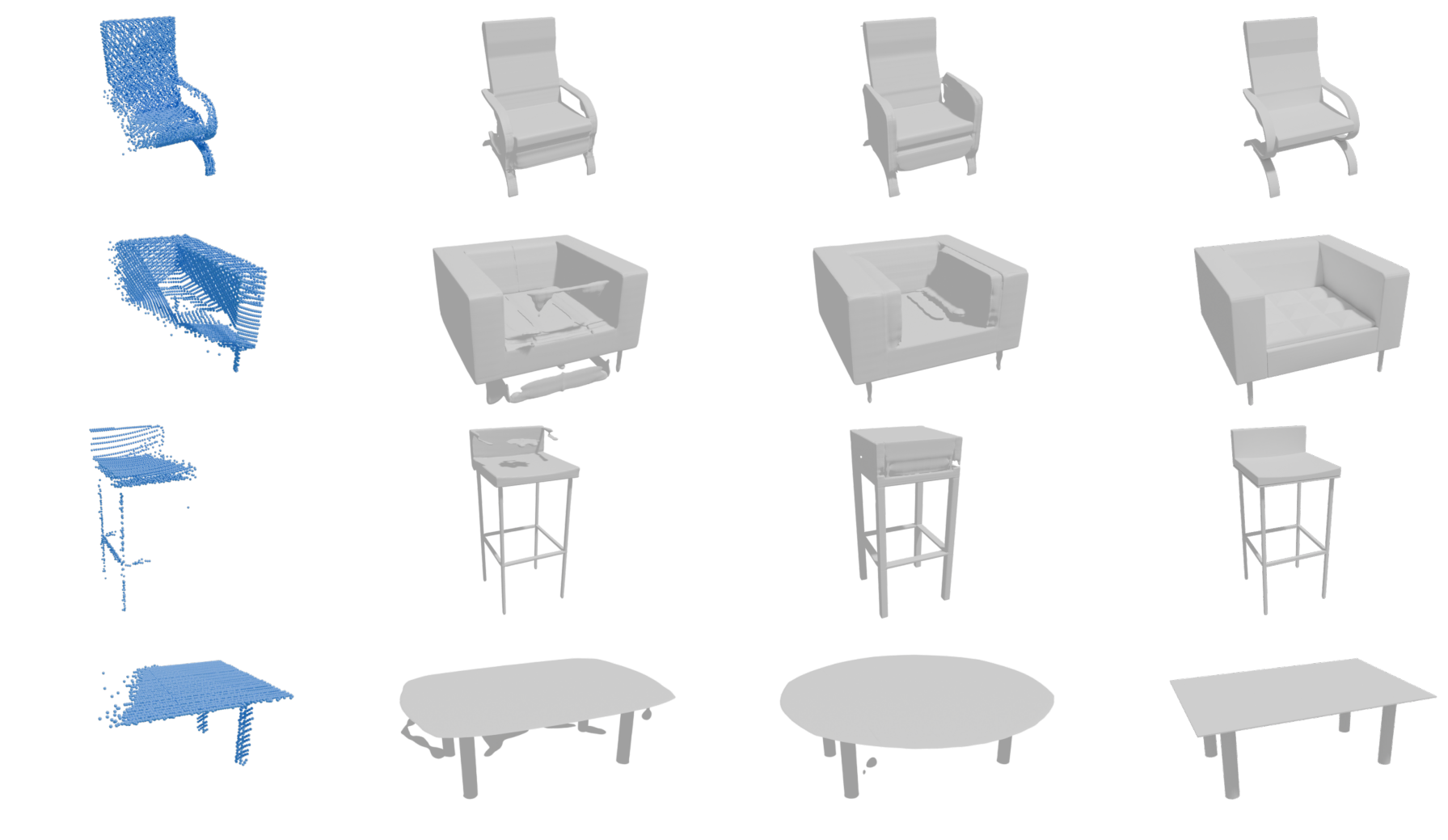}};
\foreach \x/\lab in {
    0.116279/{Input},
    0.372093/{DAPS},
    0.627907/{Ours},
    0.883721/{GT}
} {
  \path (img.north west) -- (img.north east) coordinate[pos=\x] (t);
  \node[anchor=south, text height=1.2ex, yshift=-3pt, text depth=0.25ex] at (t) {\fontsize{4.0pt}{4.0pt}\selectfont \lab};
}
\node[
  anchor=north,
  text height=1.2ex,
  text depth=0.25ex,
  yshift=2pt
] at (img.south) {\fontsize{4.0pt}{4.0pt}\selectfont Spurious geometry};
    \end{tikzpicture}
    \label{fig:failure-c2}
\end{minipage}\hfill
\vspace{-0.5cm}
\caption{\textbf{Failure cases.}
    While rare, failure cases largely fall into two categories: incomplete reconstructions, where part of the shape is missing from the reconstruction, and spurious geometry, where redundant geometry is hallucinated.
}
\label{fig:failure-cases}
\end{figure}

While GG-Langevin reconstructs the correct shape in most cases, it can get stuck in suboptimal local minima corresponding to failed reconstructions.
We find that failure cases, exemplified by \cref{fig:failure-cases}, can be broadly split into two categories.
The first category is incomplete reconstructions, where the reconstructed shape correctly fits part of the point cloud while leaving some parts incomplete.
Incomplete reconstructions can often be resolved on a per-shape basis by either increasing the number of sampling steps or increasing the noise level $\sigma$, albeit at the cost of increased runtime or a higher risk of spurious geometry.
The second category of failure cases is spurious geometry.
In this case, the estimated shape typically fits the point cloud, but chunks of geometry are added that are inconsistent with the rest of the shape.
Additional loss terms \cite{sitzmann2019siren,harenstam2024diffcd} can sometimes resolve the issue of spurious surfaces, although at the cost of a higher risk of incomplete reconstructions.

\section{Siren weight comparison}
\label{sec:siren-weight-comparison}
Our method is compatible with other loss functions at inference time without retraining.
In this section, we investigate additional regularization of our method by incorporating the Siren loss term developed by Sitzmann~\etal \cite{sitzmann2019siren},
\eq{
    \sirenloss(\latent)
    = \frac{|\Omega|\alpha}{2}\mathbb{E}_{\point\sim U(\Omega)}
    \big[
    e^{-\alpha|\dec(\latent, \point)|}
    \big],
}
with a weight factor $\mu$. 
That is, we add the term $\mu\sirenloss(\latent)$ to our geometric loss in \cref{eqn:geom_loss}.
The Siren loss has been shown to approximate the surface area of the reconstructed surface \cite{harenstam2024diffcd}, \ie $\sirenloss(\latent) \approx |S_\latent|$, where $|\cdot|$ denotes the surface area.
This means that setting $\mu > 0$ biases the reconstruction toward shapes with a lower surface area.
Consequently, using larger values of $\mu$ typically results in smoother shapes with less spurious geometry, but they also lead to more incomplete reconstructions.

\begin{figure}[t]
\centering
\begin{minipage}{0.62\textwidth}
    \centering
    \begin{tikzpicture}
    \node[inner sep=0] (img) {\includegraphics[width=\linewidth]{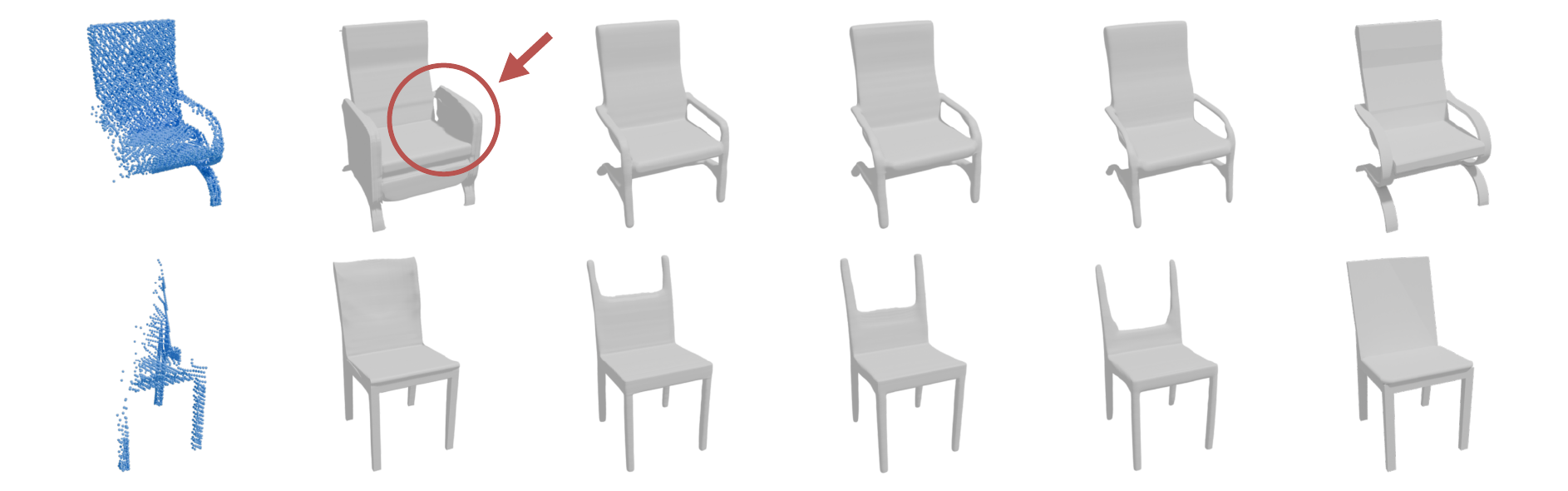}};
\foreach \x/\lab in {
    0.076923/{Input},
    0.246154/{$\mu\!=\!0$},
    0.415385/{$\mu\!=\!10^{\text{-}5}$},
    0.584615/{$\mu\!=\!10^{\text{-}4}$},
    0.753846/{$\mu\!=\!10^{\text{-}3}$},
    0.923077/{GT}
} {
  \path (img.north west) -- (img.north east) coordinate[pos=\x] (t);
  \node[anchor=south, text height=1.2ex, yshift=-3pt, text depth=0.25ex] at (t) {\fontsize{4.0pt}{4.0pt}\selectfont \lab};
}
    \end{tikzpicture}
    \label{fig:siren}
\end{minipage}\hfill
\begin{minipage}{0.32\textwidth}
    \centering
    \footnotesize
    \vspace*{\fill}
    \setlength{\tabcolsep}{3pt}
    \renewcommand{\arraystretch}{1.05}
    \resizebox{\linewidth}{!}{%
    \begin{tabular}{lcccc}
        & \multicolumn{2}{c}{\textbf{Sparse}} & \multicolumn{2}{c}{\textbf{Incomplete}} \\
        \cmidrule(lr){2-3} \cmidrule(lr){4-5}
        \textbf{Sampler} & CD & CA & CD & CA \\
        \midrule
        $\mu=10^{-3}$
	& \textbf{0.90} & \textbf{18.0} 
	& 2.63 & 23.3 
	\\
	$\mu=10^{-4}$
	& 0.97 & 18.8 
	& 1.90 & 22.3 
	\\
	$\mu=10^{-5}$
	& 0.94 & 18.4 
	& 2.02 & 23.4 
	\\
	$\mu = 0$ (Ours)
	& 0.95 & 18.6 
	& \textbf{1.41} & \textbf{18.8} 
	\\
        \bottomrule
    \end{tabular}%
    }
    \label{tab:siren}
\end{minipage}
\vspace{-0.5cm}
    \caption{\textbf{Siren weight comparison.} Increasing the Siren loss weight, $\mu$, typically leads to less spurious geometry (top row), but can also lead to incomplete reconstructions (bottom row).
    The table shows metrics averaged across all shape categories.
    }
    \label{fig:siren-weight-comparison}
\end{figure}
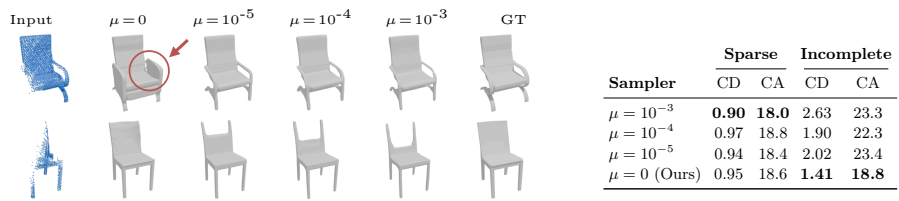

We show reconstruction results using GG-Langevin with $\mu \in \{0, 10^{-5}, 10^{-4}$, $10^{-3}\}$ in \cref{fig:siren-weight-comparison}.
Note that $\mu=0$ is our default setting, which we use for all experiments in the main manuscript.
We find that increasing $\mu$ provides a consistent benefit in the sparse setting (by smoothing out the impact of noise and reducing spurious geometry), but it also leads to consistent degradation in the performance on incomplete shapes (by removing valid parts of the geometry).
Since no setting improves both benchmarks, we set $\mu=0$ for all other experiments, noting that larger values of $\mu$ can be beneficial depending on the specific problem scenario.

\section{Limitations and future work}
As demonstrated in \cref{sec:autoencoder-ablation}, our method requires a diffusion model trained on a sufficiently well-behaved latent space. 
Existing large-scale diffusion models \cite{hunyuan3d22025tencent,li2025triposg} are typically trained with truncated SDFs, using a shallow encoder and a deep decoder.
A promising direction for future work is therefore to adapt our method to work more efficiently with a broader range of latent representations. Conversely, another direction would be to further investigate how to design latent spaces that are well-suited for the type of guidance functions that appear in surface reconstruction problems.
Our method also requires more computational time than existing feed-forward methods \cite{Huang:2023:NKS, yan2022shapeformer}.
The bulk of our computation time is spent on computing the gradients of the geometric loss.
Therefore, one highly promising extension would be to reduce the number of decoder evaluations required to reliably obtain a complete shape.
For instance, this can be achieved by adapting our method to take multiple denoising steps for each guidance step, while still maintaining measurement consistency.

\section{Class conditioning}
\label{sec:class-conditioning}

\begin{figure}[t]
\centering
\begin{minipage}{0.52\textwidth}
    \centering
    \begin{tikzpicture}
    \node[inner sep=0] (img) {\includegraphics[width=\linewidth]{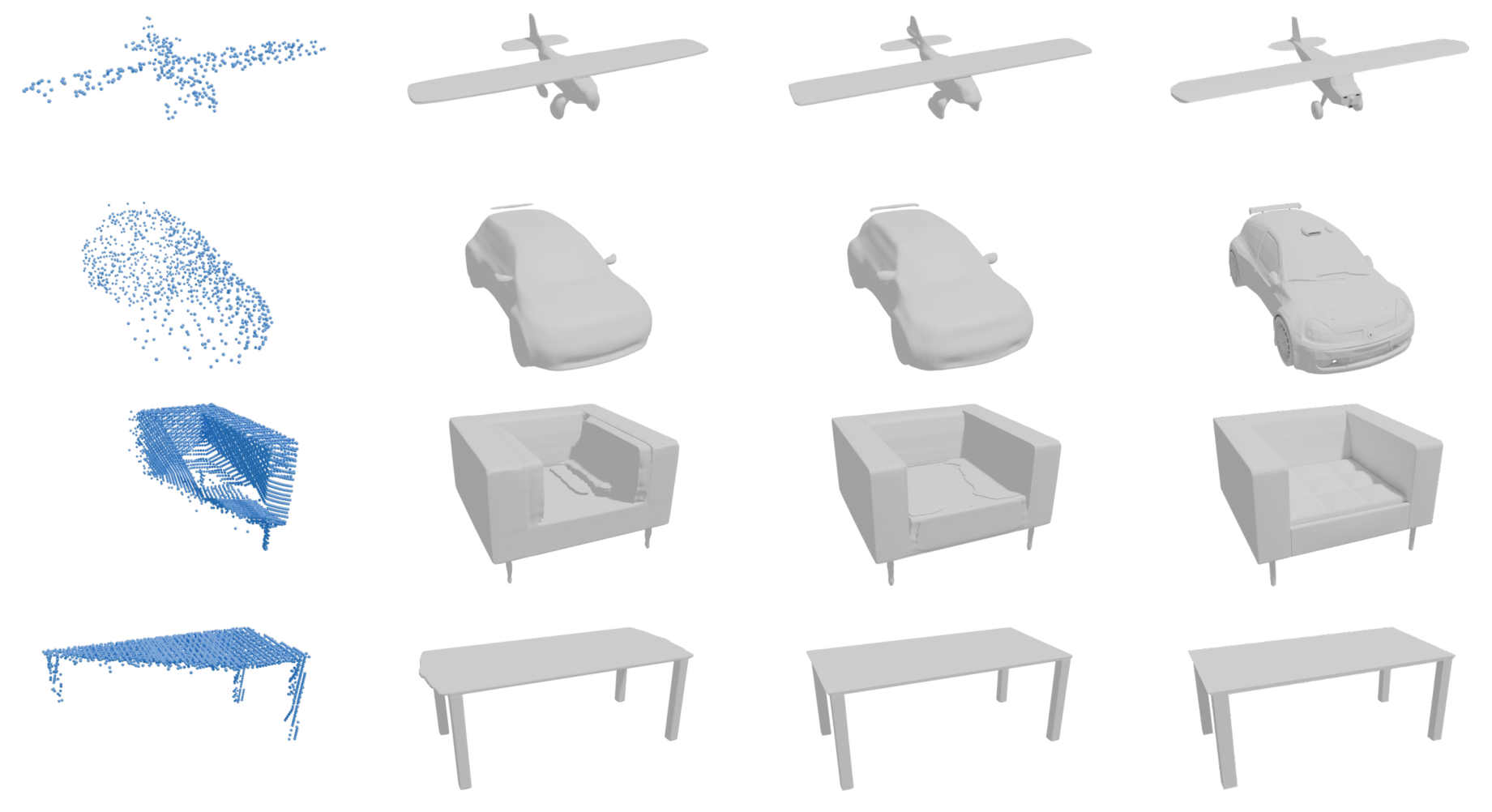}};
\foreach \x/\lab in {
    0.116279/{Input},
    0.372093/{uncond.},
    0.627907/{class-cond.},
    0.883721/{GT}
} {
  \path (img.north west) -- (img.north east) coordinate[pos=\x] (t);
  \node[anchor=south, text height=1.2ex, yshift=-3pt, text depth=0.25ex] at (t) {\fontsize{4.0pt}{4.0pt}\selectfont \lab};
}
\foreach \y/\lab in {
    0.200000/{Sparse},
    0.700000/{Incomplete}
} {
  \path (img.north west) -- (img.south west) coordinate[pos=\y] (l);
  \node[anchor=center, xshift=-2.5pt, rotate=90, text height=1.2ex, text depth=0.25ex] at (l) {\fontsize{4.0pt}{4.0pt}\selectfont \lab};
}
    \end{tikzpicture}
    \label{fig:class-cond}
\end{minipage}\hfill
\begin{minipage}{0.42\textwidth}
    \centering
    \footnotesize
    \vspace*{\fill}
    \setlength{\tabcolsep}{3pt}
    \renewcommand{\arraystretch}{1.05}
    \resizebox{\linewidth}{!}{%
    \begin{tabular}{lcccc}
        & \multicolumn{2}{c}{\textbf{Sparse}} & \multicolumn{2}{c}{\textbf{Incomplete}} \\
        \cmidrule(lr){2-3} \cmidrule(lr){4-5}
        \textbf{Sampler} & CD & CA & CD & CA \\
        \midrule
Ours (class-conditioned)
	& 0.98 & \textbf{18.6} 
	& \textbf{1.15} & \textbf{17.6} 
	\\
	Ours (unconditioned)
	& \textbf{0.95} & \textbf{18.6} 
	& 1.41 & 18.8 
	\\
        \bottomrule
    \end{tabular}%
    }
    \label{tab:class-cond}
\end{minipage}
\vspace{-0.5cm}
\caption{\textbf{Adding class conditioning.}
    Comparing GG-Langevin with and without class conditioning.
    The class-conditioned model has a clear advantage in the incomplete setting, where predicting the object category from the point cloud is more difficult.
    In the sparse setting, both models perform similarly.
    For all our main experiments, we use the unconditioned model, which does not require the class label as input.
    }
\label{fig:class-conditioning}
\end{figure}
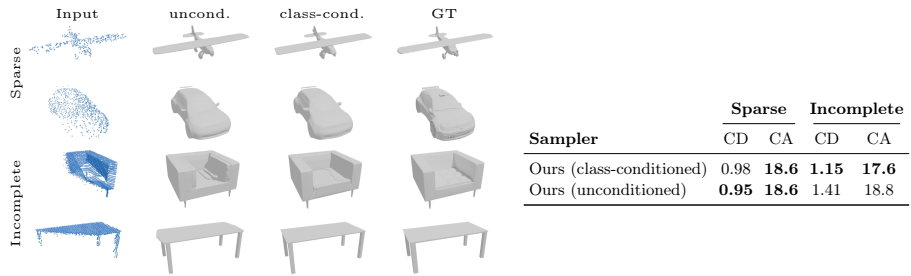

We use a diffusion model without class conditioning in all our main experiments.
This means that our method has to indirectly infer the object category (Car, Airplane, Table, or Chair) at inference time.
In order to investigate how this ambiguity impacts reconstruction performance, we also train a class-conditioned diffusion model $\epsilon_{\sigma,\theta}(\latent, c)$, which takes the object class $c$ as input.
We compare our unconditioned model with the class-conditioned model in \cref{fig:class-conditioning}.

Interestingly, we see no clear benefit in adding class conditioning in the sparse setting. In fact, in this setting, the unconditioned model achieves a slightly lower Chamfer Distance.
The benefit of class conditioning becomes more apparent in the incomplete setting.
For incomplete point clouds, our method has to rely more heavily on the prior for reconstructing the full shape. As expected, the more accurate prior leads to a more accurate reconstruction.

\section{Initializing DPS and DAPS}
\label{sec:dps-daps-adaptations}
\begin{figure}[t]
\centering
\begin{minipage}{0.52\textwidth}
    \centering
    \begin{tikzpicture}
    \node[inner sep=0] (img) {\includegraphics[width=\linewidth]{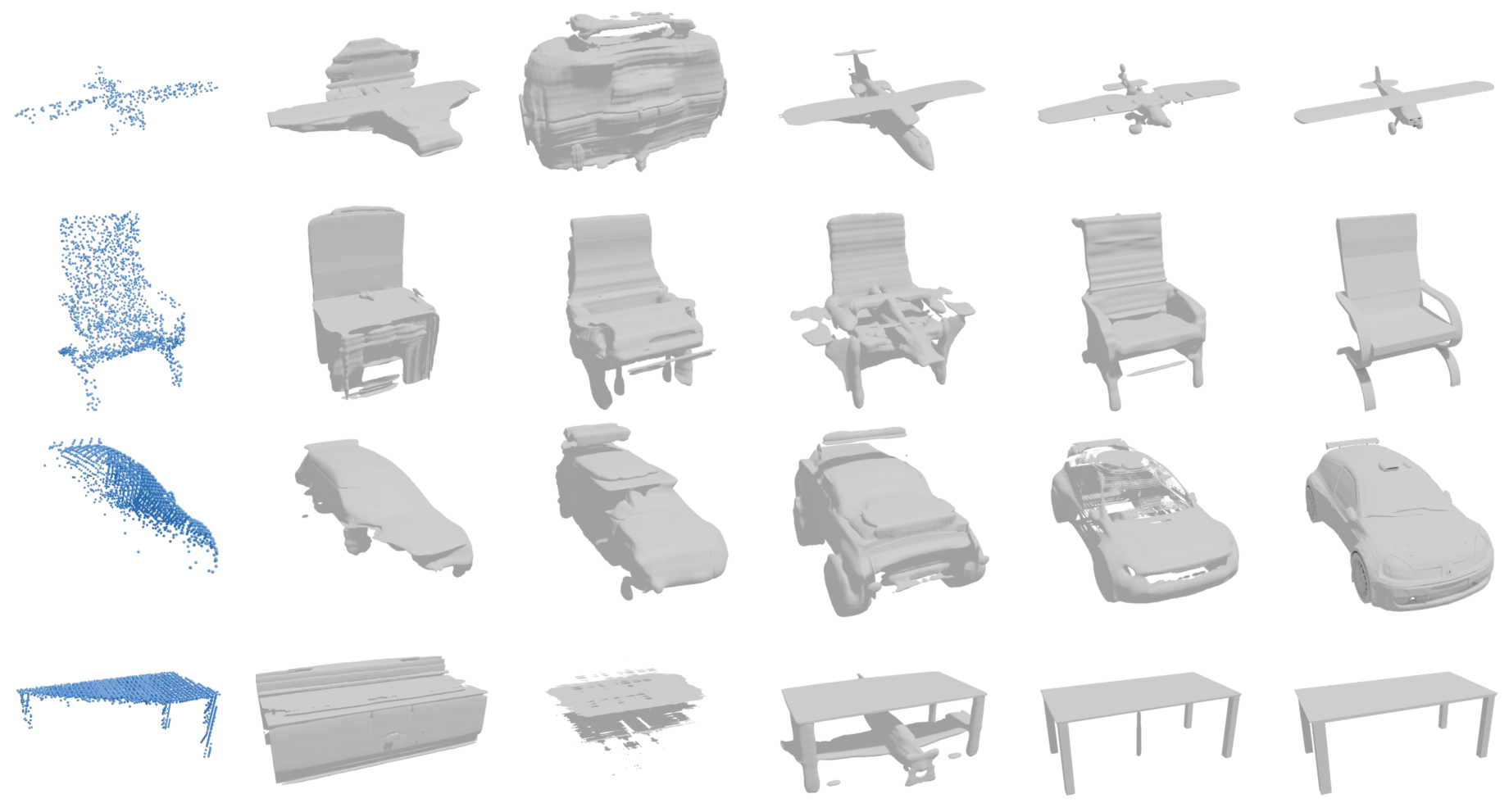}};
\foreach \x/\lab in {
   0.076923/{Input},
    0.246154/{no init},
    0.415385/{with init},
    0.584615/{no init},
    0.753846/{with init},
    0.923077/{GT}
} {
  \path (img.north west) -- (img.north east) coordinate[pos=\x] (t);
  \node[anchor=south, text height=1.2ex, yshift=-3pt, text depth=0.25ex] at (t) {\fontsize{4.0pt}{4.0pt}\selectfont \lab};
}
\foreach \x/\lab in {
    0.331/{DPS},
    0.669/{DAPS},
} {
  \path (img.north west) -- (img.north east) coordinate[pos=\x] (t);
  \node[anchor=south, text height=1.2ex, yshift=3pt, text depth=0.25ex] at (t) {\fontsize{4.0pt}{4.0pt}\selectfont \lab};
}
\foreach \y/\lab in {
    0.200000/{Sparse},
    0.700000/{Incomplete}
} {
  \path (img.north west) -- (img.south west) coordinate[pos=\y] (l);
  \node[anchor=center, xshift=-2.5pt, rotate=90, text height=1.2ex, text depth=0.25ex] at (l) {\fontsize{4.0pt}{4.0pt}\selectfont \lab};
}
    \end{tikzpicture}
    \label{fig:<LABEL>}
\end{minipage}\hfill
\begin{minipage}{0.42\textwidth}
    \centering
    \footnotesize
    \vspace*{\fill}
    \setlength{\tabcolsep}{3pt}
    \renewcommand{\arraystretch}{1.05}
    \resizebox{\linewidth}{!}{%
    \begin{tabular}{lcccc}
        & \multicolumn{2}{c}{\textbf{Sparse}} & \multicolumn{2}{c}{\textbf{Incomplete}} \\
        \cmidrule(lr){2-3} \cmidrule(lr){4-5}
        \textbf{Sampler} & CD & CA & CD & CA \\
        \midrule
DPS (no init)
	& 3.62 & 41.9 
	& 4.25 & 41.8 
	\\
	DPS (with init)
	& 3.26 & 38.7 
	& 4.04 & 37.8 
	\\
	DAPS (no init)
	& 4.33 & 36.3 
	& 4.82 & 35.9 
	\\
	DAPS (with init)
	& \textbf{1.04} & \textbf{23.2} 
	& \textbf{1.55} & \textbf{19.5} 
	\\        \bottomrule
    \end{tabular}%
    }
    \label{tab:sampler-init}
\end{minipage}
\vspace{-0.5cm}
\caption{\textbf{Initializing baselines.}
    For a fair comparison, we adapt existing methods DPS \cite{chung2022dps} and DAPS \cite{zhang2025improving} to make use of the encoder initialization $\latent_0=\enc(\pointcloud)$. In both cases, 
    our suggested
    initialization improves performance.
    }
\label{fig:sampler-initialization}
\end{figure}
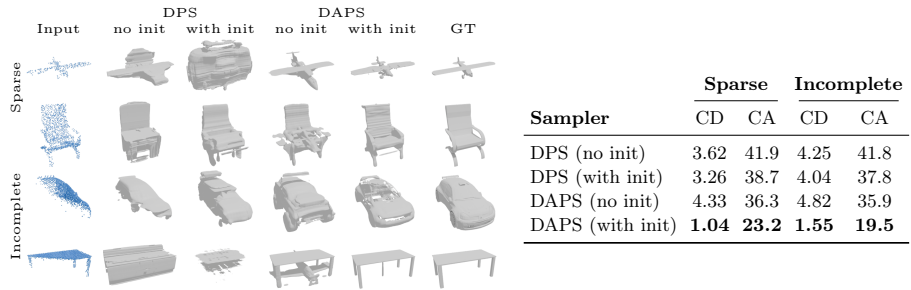

Existing methods for guided sampling like DPS \cite{chung2022dps} and DAPS \cite{zhang2025improving} start from pure noise, without any information about the point cloud.
In contrast, a key advantage of our HDND sampler is that it can be initialized directly from the VAE encoder with $\latent_0 = \enc(\pointcloud)$. 
While leveraging the initial estimate for the alternative samplers is not as straightforward, we find that both DPS \cite{chung2022dps} and DAPS \cite{zhang2025improving} can also benefit from this initial guess.
For DPS \cite{chung2022dps}, we solve the reverse SDE starting at $\latent_0 = \enc(\pointcloud)$ and ending at $\latent_0^{(T)}\sim\mathcal{N}(0,\sigma_T)$, where $\sigma_T = 80$. 
We then run the DPS \cite{chung2022dps} algorithm using $\latent_0^{(T)}$ as the initial noise vector.
For DAPS, we initialize from $\enc(\pointcloud) + \sigma_{\text{max}}n$, where $\sigma_{\text{max}} = 1$ and $n\sim\mathcal{N}(0, 1)$.
Sampling is then performed by annealing the noise level from $\sigma_{\text{max}}$ to 0 with 250 MCMC steps per noise level.
For both methods, we use the same total number of sampling steps as for GG-Langevin, resulting in the same number of calls to the diffusion model and the VAE decoder.
We show that the encoder initialization improves the performance of both DPS \cite{chung2022dps} and DAPS \cite{zhang2025improving} in \cref{fig:sampler-initialization}.

\section{Noise schedule}
\label{sec:noise-schedule-ablation}
\begin{figure}[t]
\centering
\begin{minipage}{0.72\textwidth}
    \centering
    \begin{tikzpicture}
    \node[inner sep=0] (img) {\includegraphics[width=\linewidth]{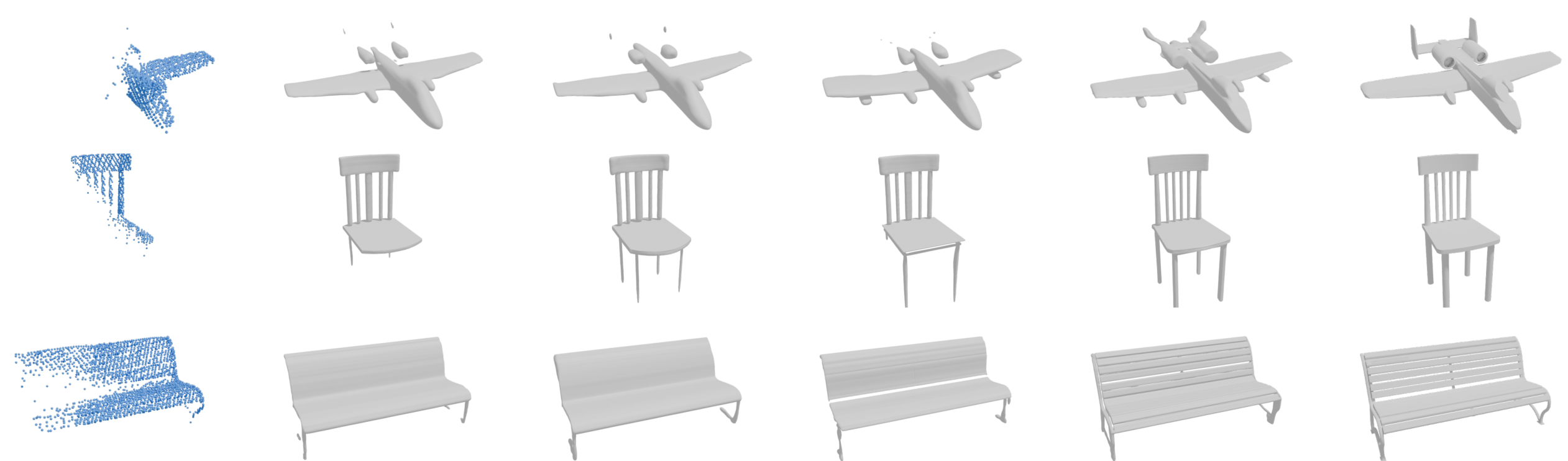}};
\foreach \x/\lab in {
    0.070354/{Input},
    0.242212/{const 2k},
    0.414071/{const 5k},
    0.585929/{cos 2k},
    0.757788/{cos 5k (Ours)},
    0.929646/{GT}
} {
  \path (img.north west) -- (img.north east) coordinate[pos=\x] (t);
  \node[anchor=south, text height=1.2ex, yshift=-3pt, text depth=0.25ex] at (t) {\fontsize{4.0pt}{4.0pt}\selectfont \lab};
}
    \end{tikzpicture}
    \label{fig:noise-sched}
\end{minipage}\hfill
\begin{minipage}{0.22\textwidth}
    \centering
    \footnotesize
    \vspace*{\fill}
    \setlength{\tabcolsep}{3pt}
    \renewcommand{\arraystretch}{1.05}
    \resizebox{\linewidth}{!}{%
    \begin{tabular}{lcc}
        \textbf{$\sigma$ schedule} & CD & CA \\
        \midrule
        const 2k
	& 1.89 & 22.5 
	\\
	const 5k
	& 1.91 & 23.3 
	\\
	cos 2k
	& 1.72 & 22.4 
	\\
	cos 5k (Ours)
	& \textbf{1.41} & \textbf{18.8} 
	\\
        \bottomrule
    \end{tabular}%
    }
    \label{tab:noise-sched}
\end{minipage}
\vspace{-0.5cm}
    \caption{\textbf{Noise schedule for incomplete scans.} Using cosine annealing for the noise level $\sigma$ (rather than constant) and increasing the number of steps from 2000 (2k) to 5000 (5k) is beneficial for incomplete point cloud scans.}
    \label{fig:noise-schedule}
\end{figure}
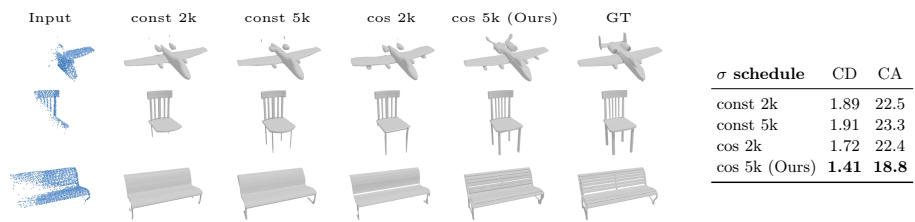

We find that the noise level parameter $\sigma$ in \cref{eqn:geom_loss} acts as a step size for the learned prior, analogous to how the guidance strength $\beta$ acts as a step size for the geometric guidance.
In this case, a single ``step'' corresponds to adding noise to the current latent $z_t$ (obtaining $\tilde z_t$), followed by half-denoising with $\frac{\sigma^2}{2}s_\sigma(\tilde \latent_t)$.
This naturally suggests applying a noise-level schedule by replacing $\sigma$ with a time-varying $\sigma_t$ to dynamically vary the trade-off between fitting the prior or fitting the measurements during sampling.

We find that scheduling the noise level is particularly important for incomplete scans.
In the incomplete setting, we need to initially rely more on the prior than on the measurements to allow the model to first complete the missing parts before focusing on details.
Thus, we find it beneficial to start the sampling in the incomplete setting with larger prior-based ``steps'', \ie large $\sigma_0 = \sigma_{\text{max}}$, and then progressively reduce the reliance on the prior throughout sampling, as the partial shape becomes increasingly complete, by decreasing $\sigma_t$ until $\sigma_N = \sigma_{\text{min}}$.
Specifically, we use a cosine-annealing schedule for the first 4000 iterations,
\eq{
\sigma_t = 
    \sigma_\text{min} + (\sigma_\text{max}-\sigma_\text{min})\cos\Big(\pi \frac{t}{4000}\Big),
}
with $\sigma_{\text{max}} = 0.2$ and $\sigma_{\text{min}} = 0.02$,
followed by $\sigma_t=\sigma_{\text{min}}$ for another 1000 iterations ($N=5000$ iterations in total).
We validate these design choices with an experiment in \cref{fig:noise-schedule}.
The default setting with constant $\sigma = 0.05$ for $N=2000$ iterations (const 2k) achieves relatively poor performance, since the shape remains incomplete at the end of sampling.
We find it optimal to combine cosine scheduling while increasing the number of steps to $N=5000$ (cos 5k).

Note that the $\sigma$-schedule in GG-Langevin is distinct from the noise level schedules typically used for reverse SDE sampling \cite{song2020score,karras2022elucidating}.
Notably, we do not require a large initial noise level $\sigma_0\gg 1$, that the noise level must tend toward zero, $\sigma_N \approx 0$, or that $\sigma_t$ is a decreasing function.
Our only assumptions about $\sigma_t$ are that it is sufficiently small such that the half-denoising approximation applies \cite{hyvarinen2024noise}, and that it is constant towards the end of sampling, such that Langevin dynamics converges to a stable distribution.
Our final stage with 1k steps at a constant noise level $\sigma=0.02$ satisfy both of these criteria.

\end{document}